\documentclass[letterpaper]{article} 
\usepackage[submission]{aaai25}  
\usepackage{comment}
\usepackage{times}  
\usepackage{helvet}  
\usepackage{courier}  
\usepackage[hyphens]{url}  
\usepackage{graphicx} 
\usepackage{natbib}  
\usepackage{caption} 
\usepackage{amsmath} 
\usepackage{siunitx}
\usepackage[linesnumbered,ruled,vlined]{algorithm2e}
\usepackage[dvipsnames]{xcolor} 
\definecolor{dark-blue}{rgb}{0,0,0.55}
\definecolor{d-blue}{rgb}{0,0,0.7}
\definecolor{rw}{rgb}{0.5,0.5,0.5}
\definecolor{rwp}{rgb}{0,0,1}
\newcommand{\rw}[1]{\textcolor{rw}{#1}}
\newcommand{\rwp}[1]{\textcolor{rwp}{#1}}
\usepackage{cite} 
\usepackage{colortbl}        
\definecolor{Red}{rgb}{0.933, 0.49, 0.42}  
\definecolor{cyan}{rgb}{0.42, 0.78, 0.95}  
\usepackage{url}            
\usepackage{booktabs}       
\usepackage{amsfonts}       
\usepackage{nicefrac}       
\usepackage{microtype}      
\usepackage{multirow}
\usepackage{makecell}

\usepackage{newfloat}
\usepackage{listings}
\DeclareCaptionStyle{ruled}{labelfont=normalfont,labelsep=colon,strut=off} 
\title{Reevaluation of Large Neighborhood Search for MAPF: \\Findings and Opportunities}

\author{
    Jiaqi Tan\textsuperscript{\rm 1}\equalcontrib,
    Yudong Luo\textsuperscript{\rm 2}\equalcontrib,
    Jiaoyang Li\textsuperscript{\rm 3},
    Hang Ma\textsuperscript{\rm 1}
}
\affiliations{
    \textsuperscript{\rm 1}Simon Fraser University,
    \textsuperscript{\rm 2}University of Waterloo,
    \textsuperscript{\rm 3}Carnegie Mellon University\\
    \{jiaqit, hangma\}@sfu.ca,
    yudong.luo@uwaterloo.ca,
    jiaoyangli@cmu.edu
}

\usepackage{bibentry}

\begin{document}

\maketitle

\begin{abstract}

Multi-Agent Path Finding (MAPF) aims to arrange collision-free goal-reaching paths for a group of agents. Anytime MAPF solvers based on large neighborhood search (LNS) have gained prominence recently due to their flexibility and scalability, leading to a surge of methods, especially those leveraging machine learning, to enhance neighborhood selection. However, several pitfalls exist and hinder a comprehensive evaluation of these new methods, which mainly include: 1) Lower than actual or incorrect baseline performance; 2) Lack of a unified evaluation setting and criterion; 3) Lack of a codebase or executable model for supervised learning methods. To address these challenges, we introduce a unified evaluation framework, implement prior methods, and conduct an extensive comparison of prominent methods. Our evaluation reveals that rule-based heuristics serve as strong baselines, while current learning-based methods show no clear advantage on time efficiency or improvement capacity. Our extensive analysis also opens up new research opportunities for improving MAPF-LNS, such as targeting high-delayed agents, applying contextual algorithms, optimizing replan order and neighborhood size, where machine learning can potentially be integrated. Code and data are available at \url{https://github.com/ChristinaTan0704/mapf-lns-unified}.

\end{abstract}

\section{Introduction}
\label{sec:intro}

Multi-Agent Path Finding (MAPF) refers to the problem of arranging collision-free paths for a group of agents~\cite{stern2019multi}. Many real-world applications involving multiple agents are closely related to MAPF, e.g., warehouse robots~\cite{ma2017ai,li2021lifelong}, aircraft-towing vehicles~\cite{morris2016planning,fines2020agent}, and navigation in video games~\cite{ma2017feasibility}. 

MAPF is NP-hard to solve optimally~\cite{yu2013structure}. In recent years, anytime MAPF solvers based on large neighborhood search (LNS)~\cite{li2021anytime} have gained prominence since previous centralized solvers often suffer from poor efficiency with low scalability despite their solution optimality, e.g, conflict-based search (CBS)~\cite{sharon2015conflict}, or low solution quality despite their fast speed and good scalability, e.g., prioritized planning (PP)~\cite{erdmann1987multiple}. Learning decentralized suboptimal policies via reinforcement learning has also been explored, but typically requires subtle environment design~\cite{sartoretti2019primal,ma2021distributed,ma2021learning}. Among these approaches, MAPF-LNS has emerged as the leading method for finding fast and near-optimal solutions to large-scale MAPF problems within a time budget. It starts from a fast initial solution, often obtained using a fast suboptimal MAPF solver, e.g., PP, and incrementally improves the solution quality to near-optimal over time. In LNS, a subset of agents, called a neighborhood, is selected, and their paths are iteratively destroyed and repaired. MAPF-LNS has consistently ranked first in various competitions, including the 2023 Robot Runners~\cite{jiang2024scaling}, AMLD 2021 Flatland 3 Challenge~\cite{chen2023scalable}, and the 2020 NIPS Flatland Challenge~\cite{li2021scalable}, demonstrating its excellence in both speed and solution quality.

One key challenge of MAPF-LNS lies in selecting an improving neighborhood to efficiently minimize total delays.
To address this, various strategies have been proposed, which generally fall into two categories: \textbf{rule-based} and \textbf{learning-based} methods. Rule-based strategies rely on predefined heuristics to generate neighborhoods~\cite{li2021anytime}, while learning-based strategies predict the optimal neighborhood generated by rule-based strategies~\cite{huang2022anytime, yan2024neural} or dynamically select one of the rule-based strategies based on environmental conditions~\cite{phan2024adaptive}. However, as an emerging research topic, no current work systematically examines the efficiency of different MAPF-LNS methods, especially the new advances using machine learning. Upon examination, we find several pitfalls in their evaluation, which impede a reasonable comparison. This includes: 1) \textbf{Underreported or incorrect performance}. We observe that the final delays of rule-based methods reported in~\citet{huang2022anytime} are usually significantly higher than those in~\citet{li2021anytime}. \citet{phan2024adaptive} directly import these values from~\citet{huang2022anytime}. Additionally,~\citet{yan2024neural} adopts final delays from~\citet{li2021anytime} but uses slightly different maps for evaluation, leading to inconsistencies. For example, the result of map `random-32-32-20' in~\citet{li2021anytime} has been incorrectly adopted as the result for `random-32-32-10' by ~\citet{yan2024neural}. Such discrepancies make it difficult to draw reliable conclusions from current results. 2) \textbf{Lack of a unified setting}. Various factors potentially influence the efficiency of MAPF-LNS, such as the initial solution and path replan solver. Initial solutions generally vary among different MAPF-LNS papers, and unsolved scenarios are usually discarded. While PP serves as the default replan solver in most MAPF-LNS methods, priority-based search (PBS)~\cite{ma2019searching} is used by~\citet{yan2024neural}. Different evaluation metrics are also employed in different papers, such as area under the curve (AUC)~\cite{li2021anytime}, win/loss~\cite{huang2022anytime}, and average gap~\cite{yan2024neural}. This lack of a unified setting makes direct comparison difficult. 3) \textbf{Lack of a codebase or executable model}. The performance of supervised learning heavily depends on data quality and parameter tuning. However, codebases, running instructions, or executable models are generally missing for supervised learning methods, making it challenging to reproduce their results.

To address these issues, we propose a unified evaluation setting under the same benchmark and hyperparameter configurations. We investigate and standardize several key aspects in MAPF-LNS, including initial solutions, replan solvers, and time-counting schemes during evaluation, which are not fully studied, obscure, or incorrect in previous works. We then implement and reevaluate prior methods in this unified framework. Our key finding is that rule-based heuristics for neighborhood selection are still strong baselines compared to learning-based methods in terms of time efficiency and improvement capacity. Our analysis also leads to several interesting future directions for improving MAPF-LNS, which are less explored in the previous MAPF-LNS literature, e.g., targeting high-delayed agents, applying contextual algorithms, optimizing replan order and neighborhood size, where machine learning can potentially be integrated.


\section{Preliminaries}
\subsection{Background: MAPF and MAPF-LNS}

The MAPF variants are summarized by~\citet{stern2019multi}. In this work, we follow the common settings: 1) considering vertex and swapping conflicts, i.e., agents can not occupy the same vertex or traverse the same edge in opposite directions simultaneously; 2) agents `stay at target' instead of disappearing; 3) minimizing the sum of individual costs, i.e., the total time steps for all agents to reach their targets. 

\textbf{MAPF} is formally defined as follows. The input is a graph $G=(V,E)$, where $V$ is the set of vertices and $E$ is the set of edges, along with a set of $N$ agents $A=\{a_1,...,a_N\}$. 
Each agent $a_i$ is assigned a start vertex $s_i\in V$ and a goal (target) vertex $g_i\in V$ ($g_i$ is accessible from $s_i$). At each discrete time step, an agent can either \textit{move} to an adjacent vertex or \textit{wait} at its current vertex. Consequently, the path $p_i$ of $a_i$ consists of a sequence of vertices that are adjacent or identical. A \textit{solution} is a set of collision-free paths, one for each agent from $s_i$ to $g_i$. Let $d(s_i,g_i)$ denote the length of the shortest path between $s_i$ and $g_i$, and $l(p_i)$ denote the length of path $p_i$. Then, the delay of path $p_i$ is  $delay(p_i)=l(p_i)-d(s_i,g_i)$. Note that $l(p_i)$ counts the edges of both \textit{move} and \textit{wait} actions. The task is to find a \textit{solution} $P=\{p_i\}_{i=1}^N$ that minimizes the sum of costs $\sum_{i=1}^N l(p_i)$, which equals to minimize the sum of delays $\sum_{i=1}^N delay(p_i)$.

\textbf{LNS} is a type of improvement heuristic that iteratively reoptimizes a solution by the \textit{destroy} and \textit{repair} operations until some stopping condition is met~\cite{pisinger2010large}. In the \textit{destroy} operation, it breaks a part of the solutions, named a \textit{neighborhood}. In \textit{repair} operation, it solves the reduced problem by treating the remaining part as fixed. 

\textbf{MAPF-LNS} framework operates as follows: given a MAPF instance, an initial (non-optimal) \textit{solution} $P_0$ is obtained via a non-optimal solver. In each iteration $k$, a subset of agents $\tilde{A}\subset A$ is selected based on a criterion. $\tilde{A}$ is called a {neighborhood}. The paths of agents in $\tilde{A}$ are then removed from  previous {solution} $P_{k-1}$, resulting in $P^-_{k-1}=\{p_i\in P_{k-1}| a_i\notin \tilde{A}\}$. Subsequently, those paths of $\tilde{A}$ are replanned by an algorithm to avoid collisions with each other and with the paths in $P^-_{k-1}$. If the resulting {solution} has a smaller sum of delays than $P_{k-1}$, it is accepted as $P_k$, otherwise, $P_k$ remains as $P_{k-1}$.

\subsection{Neighborhood Selection }\label{sec:heuristics}
Selecting the neighborhood is crucial to the success of MAPF-LNS.  In this section, we give an overview of the existing selection strategies. 

\textbf{Rule-based strategies:}
There are four major rule-based strategies in the literature: \textbf{RandomWalk}, \textbf{Intersection}, \textbf{Random}, and \textbf{Adaptive}, proposed by~\citet{li2021anytime}. 
Different rule-based strategies improve the current solution from different perspectives. \textit{RandomWalk} lets an unoptimized high-delayed agent move towards a shorter path and collect colliding agents and itself as a neighborhood. \textit{Intersection} focuses on improving solutions around intersection vertices (vertices with a degree greater than two) by adding agents that visit the same intersection to the neighborhood.
\textit{Random} strategy selects agents uniformly at random from the set of agents, ensuring a broad exploration of the solution space. Although simple, this method introduces sufficient diversity, preventing the algorithm from getting stuck in local optima, and is widely used in LNS~\cite{demir2012adaptive, song2020general}. 
\textit{Adaptive} dynamically switches between RandomWalk, Intersection, and Random strategies, adjusting their sampling probability weights based on relative success in improving the current solution.

\textbf{Learning-based methods:} There are three 
 prominent learning-based strategies: \textbf{SVM-LNS}~\citep{huang2022anytime} (denoted as SVM), \textbf{Neural-LNS}~\citep{yan2024neural} (denoted as NNS), and \textbf{Bandit-LNS}~\citep{phan2024adaptive} (denoted as Bandit). \textit{SVM} and \textit{NNS} are supervised learning methods, where a ranking model is trained to predict the best neighborhood over a set of neighborhood candidates generated by rule-based strategies. \textit{Bandit} incorporates a bandit algorithm to select rule-based strategies and neighborhood size as bi-level arms. The reward signal is the delay improvement.

\subsection{Discussion on  Selection Strategies}\label{sec:discussion-learning}
\textbf{RandomWalk}~\citep{li2021anytime}. The main idea of this heuristic is to prioritize replanning for agents with high delays. On investigating this method, we find some designs of its algorithm may hinder re-optimizing high-delayed agents, e.g., once a high-delayed agent is selected in one round, it is added to a tracking list that will not be selected in next round. Also, agents are randomly chosen if the neighborhood size is not reached after one RandomWalk search. We make two modifications to RandomWalk by removing the tracking list and sampling agents according to their delays. We name this modified heuristic as \textbf{RandomWalkProb}, and we find these simple modifications lead to significant improvement in several domains. (See Sec.~\ref{sec:randomwalkprob} in Appendix on details of RandomWalkProb.)

\textbf{SVM}~\citep{huang2022anytime} and \textbf{ NNS}~\citep{yan2024neural}. Both two utilize supervised learning to rank neighborhood candidates given the query information, i.e., some solution $P_k$ in iteration $k$. This implies the queries $\{P_k\}_{k=1}^T$ are treated as independent and identically distributed (i.i.d.) during training. However, they are non i.i.d. in practice since $P_{k+1}$ highly depends on $P_{k}$ within the sequential optimization of LNS. In the literature of learning to rank, several works suggest explicitly capturing the temporal information among queries to increase the robustness and generalization ability of rank models~\citep{yu2019multi,li2020time}.

\textbf{Bandit}~\citep{phan2024adaptive}. MAPF-LNS can be framed as a contextual bandit problem, where optimal actions (e.g., strategies and neighborhood sizes) should be determined based on the context information (e.g., the map and the solution $P_k$ at iteration $k$). Different contexts can be treated as distinct states. However, Bandit-LNS employs non-contextual bandit algorithms, which operate under the assumption of a single state or no state at all, and thus select optimal actions without considering contextual information. This creates a theoretical inconsistency, as the problem setting (contextual) does not align with the algorithm applied (non-contextual).

\section{A Unified Setting for Evaluation}\label{sec:unify-setting}
In this section, we elaborate our unified setting for MAPF-LNS evaluation. We investigate and standardize several key aspects of MAPF-LNS before conducting a comprehensive comparison among existing methods.

\subsection{Environments} 
MAPF algorithms are generally evaluated on MAPF benchmark suite\footnote{\vspace{-0.4cm}https://movingai.com/benchmarks/mapf/index.html}, which provides 2D grid maps of different layouts simulating various real-world environments, such as warehouses and empty rooms. 
Among many maps in the suite, we choose six representative maps (i.e., commonly chosen maps in the aforementioned papers and cover diverse layouts) from each MAPF benchmark category: \underline{empty-32-32} of size $32\times 32$ (denoted as empty), \underline{random-32-32-20} of size $32\times 32$ (denoted as random), \underline{warehouse-10-20-10-2-1} of size $161\times 63$ (denoted as warehouse), \underline{ost003d} of size $194\times 194$, \underline{den520d} of size $256\times 257$, and \underline{Paris\_1\_256} (denoted as Paris) of size $256\times 256$. We utilize the '25 random scenarios' included in the suite, where each scenario offers a distinct set of agent start and goal locations for a given map and specified number of agents. For methods requiring training data, e.g., SVM-LNS and Neural-LNS, we generate additional scenarios using the same map layouts with new random start-goal pairs for model training, such that all methods are evaluated on the same 25 scenarios in the benchmark.

\subsection{Initial Solution} 
As an anytime algorithm, we would expect MAPF-LNS to quickly find an initial feasible solution and then improve its quality to near-optimal as time progresses. Therefore, we set the time limit for finding the initial solution to $10$ seconds by following~\citet{li2021anytime}.

Three representative suboptimal MAPF solvers, discussed by~\citet{li2021anytime}, are considered as potential initial solvers: Explicit Estimation CBS (EECBS)~\cite{li2021eecbs}, Prioritized Planning (PP)~\cite{erdmann1987multiple} with a random priority ordering, and Parallel Push and Swap (PPS)~\cite{sajid2012multi}. However, as highlighted by~\citet{li2021anytime} (see Fig 3 of~\citet{li2021anytime}), none of the three successfully solve all 25 scenarios across varying numbers of agents within the time limit, yielding unsolved scenes being discarded. To ensure that initial solutions are available for all 25 scenarios within $10$s, we adopt the following two methods as initial solvers.

1) \underline{LNS2}~\cite{li2022mapf}. It is an improved version of PP. It repeatedly repairs the collisions met by PP until the paths become collision-free. 2) \underline{LaCAM2}~\cite{okumura2023lacam2}. It was recently proposed as a fast suboptimal MAPF method. Though it is faster than LNS2, its solution quality is generally worse than LNS2.

\begin{table}[b!]
\setlength{\tabcolsep}{3.5pt}
\vspace{-12pt}
    \caption{Final delay and total iteration of using PP and PBS within $60$s when initial solver is LNS2 neighborhood selection strategy is RandomWalk and neighborhood size is 25. Cases where PBS performs better are highlighted in \textcolor{red}{red}.}
\vspace{-8pt}
    \label{table:pp-pbs-60s}
    \centering

    \resizebox{0.485\textwidth}{!}{
    \begin{tabular}{@{}c|c|S[table-format=5.1]r|cc|c|c|rr|cc@{}}
    \toprule
    \multicolumn{12}{c}{Initial Solver: LNS2; Time limit: 60s} \\
    \toprule
    \multirow{2}{*}{} & 
    \multirow{2}{*}{N} & \multicolumn{2}{c|}{Final Delay} & \multicolumn{2}{c|} {Iter (x1k)} & \multirow{2}{*}{} & \multirow{2}{*}{N} & \multicolumn{2}{c|}{Final Delay} & \multicolumn{2}{c}{Iter (x1k)} \\
     \cmidrule(lr){3-6} \cmidrule(l){9-12} 
    & & PP & \multicolumn{1}{c|}{PBS} & PP & PBS & & & \multicolumn{1}{c}{PP} &  \multicolumn{1}{c|}{PBS} & PP & PBS \\
    \midrule
    \multirow{5}{*}{\rotatebox{90}{empty}} 
    & 300 & 431.7 & 436.9 & 8.98 & 0.44 & \multirow{5}{*}{\rotatebox{90}{random}} & 150 & 350.1 & \textcolor{red}{346.9} & 7.39 & 0.63 \\
    & 350 & 1109.8 & \textcolor{red}{1081.8}& 4.22 & 0.25 & & 200 & 959.6 & \textcolor{red}{875.5} & 2.88 & 0.19 \\
    & 400 & 2570.1 & \textcolor{red}{2238.2} & 2.28 & 0.15 & & 250 & 2423.8 & \textcolor{red}{2301.4} & 1.99 & 0.05 \\
    & 450 & 4873.5 & \textcolor{red}{4293.6} & 1.83 & 0.09 & & 300 & 5309.6 & \textcolor{red}{4533.1} & 1.47 & 0.03 \\
    & 500 & 7817.6 & \textcolor{red}{6874.2} & 1.51 & 0.05 & & 350 & 8966.9 & \textcolor{red}{8076.5} & 1.57 & 0.02 \\
    \midrule
    \multirow{5}{*}{\rotatebox{90}{warehouse}} 
    & 150 & 122.1 & 128.3 & 6.59 & 0.57 & \multirow{5}{*}{\rotatebox{90}{ost003d}} & 200 & 183.9 & 897.6 & 1.75 & 0.06 \\
    & 200 & 266.8 & 310.2 & 2.65 & 0.27 & & 300 & 915.5 & 4630.9 & 0.92 & 0.02 \\
    & 250 & 477.6 & 760.3 & 1.83 & 0.15 & & 400 & 3230.7 & 8032.7 & 0.52 & 0.03 \\
    & 300 & 832.7 & 1740.3 & 1.15 & 0.09 & & 500 & 9335.3 & 16709.3 & 0.21 & 0.01 \\
    & 350 & 1495.0 & 3237.5 & 0.73 & 0.06 & & 600 & 17998.3 & 24525.7 & 0.15 & 0.01 \\
    \midrule
    \multirow{5}{*}{\rotatebox{90}{den520d}} 
    & 500 & 899.6 & 6195.8 & 1.28 & 0.05 & \multirow{5}{*}{\rotatebox{90}{Paris}} & 350 & 82.2 & 383.7 & 5.98 & 0.17 \\
    & 600 & 1321.3 & 8485.5 & 1.72 & 0.06 & & 450 & 136.5 & 2274.2 & 6.44 & 0.11 \\
    & 700 & 4436.5 & 16642.9 & 0.78 & 0.02 & & 550 & 219.3 & 4878.6 & 4.72 & 0.06 \\
    & 800 & 7342.8 & 21909.0 & 0.61 & 0.02 & & 650 & 317.1 & 9304.6 & 4.49 & 0.04 \\
    & 900 & 13032.0 & 29352.2 & 0.44 & 0.01 & & 750 & 614.9 & 14707.1 & 3.07 & 0.03 \\
    \bottomrule
    \end{tabular}
    }

\vspace{1pt}
\parbox{0.48\textwidth}{\scriptsize Note: `N' is the number of agents.}
\vspace{-10pt}
\end{table}

\subsection{Replan Solver} \label{sec:pp-vs-pbs}
The replan solver is invoked iteratively to update paths for the neighborhood and refine the solution in real-time, making a fast and efficient solver highly desirable. Except for~\citet{yan2024neural}, most methods typically select PP as the replan solver due to its fast speed in completing a single iteration (e.g., faster than CBS and EECBS~\cite{li2021anytime}), allowing for more iterations within the time limit. However, \citet{yan2024neural} argues that while Priority-Based Search (PBS) is more time-consuming per iteration, it can outperform PP in certain scenarios due to its greater improvement in a single iteration. To evaluate the efficiency of PP and PBS, we set the neighborhood size to 25 and the time limit to $60$s (neighborhood size 25 is recommended in 3/5 cases by \citet{yan2024neural}, and $60$s is used in their plots). The time-counting criterion for evaluation is detailed in Sec.~\ref{sec:eval-time-count}. The initial solver used is LNS2, with RandomWalk as the heuristic.

The final delays in different maps with various amount of agents by using PP and PBS are shown in Table~\ref{table:pp-pbs-60s}. Though our results coincide with \citet{yan2024neural} that PBS is better in empty and random maps, it is significantly worse than PP in larger maps with more agents. We also include the number of iterations performed by PP and PBS in the table. We see that PP runs notably faster than PBS and thus can explore a substantially larger number of neighborhoods within the time limit. Therefore, we choose PP as the replan solver. (More comparison results between PP and PBS are shown in Table~\ref{table:full-pp-pbs-300s-60s} in Appendix, which include cases where the initial solver is LaCAM2, and the time limit is $300$s. These results also suggest that PP is better than PBS in most cases.)

\subsection{Neighborhood Size and Number of Agents} 
Intuitively, a smaller neighborhood size accelerates each iteration but may yield limited improvements, whereas a larger neighborhood size slows down each iteration but has the potential to deliver more significant improvements. To evaluate those strategies, we test a variety of neighborhood sizes, i.e., $\{2,4,8,16,32\}$, by following~\citet{li2021anytime}.  
 
The number of agents in a map affects the congestion level of a MAPF problem. We select a broad spectrum of agent amounts in each map to encompass the range of numbers evaluated in previous papers. The number of agents evaluated for each map is summarized in Table~\ref{table:diff-eval-setting} in Appendix.

\subsection{Evaluation Criterion}\label{sec:eval-time-count}
Given the time-sensitive nature of MAPF-LNS, we mainly focus on the relationship between delay and time. Specifically, we report the \textbf{final delay} and \textbf{area under the curve} (AUC) of the delay-versus-time curve within a specified time limit. A common criterion is a time limit of $60$s. However, we observe that the delay may not converge within $60$s. To address this, we extend the time limit to $300$s (we still report the results when the time limit is $60$s). To reduce the influence of overhead from other operations, we only measure the time spent on the core processes of each method, i.e., the time used for destroying (remove agents) and repairing (replan paths). Note that SVN-LNS and Neural-LNS require calling the rule-based methods to propose neighborhood candidates and calling the trained model to predict the top neighborhood. This part of time is included in the time used by these two methods. The model prediction of Neural-LNS is performed on GPU to accelerate.

\subsection{Implementation Details}\label{sec:implementation}
We develop based on the codebase of~\citet{li2021anytime} to produce results for rule-based strategies, including RandomWalk, Intersection, Random, Adaptive, and RandomWalkProb. We use the codebase of~\citet{phan2024adaptive} to produce results for Bandit, where Thompson Sampling is the underlying bandit algorithm since it performs the best in our experiments. 
No open-source codebase is available for SVM, and neither executable models nor training data are provided for NNS. As a result, we implement and train both methods. Note that the original SVM and NNS use different initial solutions for different maps and are not clearly specified. We fix the initial solution to LNS2~\cite{li2022mapf} when recovering their results. Please refer to Sec.~\ref{sec:svm-nns-train} in Appendix for training details, e.g., dataset construction and hyperparameter tuning.

\begin{table}[t]
\caption{Reproduced and reported metrics of SVM and NNS.} 
\label{tab:validation-svm-nns}
\vspace{-6pt}
\setlength{\tabcolsep}{3pt} 
\resizebox{0.47\textwidth}{!}{%
\begin{tabular}{@{}c|c|cccc|c|S[table-format=6.1]S[table-format=6.0]|S[table-format=4.1]S[table-format=4.0]S[table-format=6.1]S[table-format=6.0]@{}}
\toprule
\multicolumn{4}{c}{\textbf{Average Rank of SVM}} & \multirow{7}{*}{\rule{0.7pt}{9.7\normalbaselineskip}} & \multicolumn{5}{c}{\textbf{Reduction in delays of NNS}} \\ 
\midrule

%

\multicolumn{1}{c}{\textbf{}} & \multicolumn{1}{c|}{\textbf{}} & \multicolumn{2}{c}{\textbf{Rank / Total}} &  & \multicolumn{1}{c}{\textbf{}} & \multicolumn{1}{c|}{\textbf{}} & \multicolumn{2}{c|}{\textbf{Initial delays}} & \multicolumn{2}{c}{\textbf{Delay elimination}} \\ 

\midrule

\textbf{Map} & \textbf{N} & \textbf{Reproduce} & \textbf{Report} &  & \textbf{Map} & \textbf{N} & \textbf{Reproduce} & \textbf{Report} & \textbf{Reproduce} & \textbf{Report} \\ \midrule
warehouse & 100 & 5.9/20 & 6.0/20 &  & empty & 300 & 3140.7 & 3200 & 2202 & 1850 \\ \midrule
ostd003d & 100 & 6.2/20 & 5.4/20 &  & warehouse & 100 & 5731.8 & 5500 & 4908 & 4700 \\ \midrule
den520d & 200 & 7.6/20 & 7.0/20 &  & ost003d & 100 & 8690.4 & 8600 & 4585.1 & 5000 \\ \midrule
Paris & 250 & 7.7/20 & 6.7/20 &  & den520d & 200 & 20510.2 & 20500 & 12721.3 & 14500 \\ \bottomrule
\end{tabular}%
}

\vspace{1pt} 
\parbox{0.47\textwidth}{\scriptsize Note: `N' is the number of agents. `Reproduce' is our reproduced results. `Report' 
is reported values of SVM~\cite{huang2022anytime} and NNS~\cite{yan2024neural}. 
}
\vspace{-9pt}
\end{table}

To validate the reliability of our reproductions of SVM and NNS, we first conduct the following sanity checks. \textit{1) Matching the reported statistics.} We reproduce SVM and NNS by training models using instructions and parameters provided in their original papers. The time measurement scheme is not clearly detailed in ~\citet{huang2022anytime,yan2024neural}. Thus, we compare the delay-versus-iteration of our reproduced results with reported ones, which is independent of hardware specifications and time-counting schemes. For NNS, we compare the reduction in delays after 100 iterations, starting from roughly the same initial delays in shared maps (we use PBS as the replan solver in this comparison for consistency with~\citet{yan2024neural}). The results of our trained NNS and reported performances are shown in the right half of Table~\ref{tab:validation-svm-nns}. The maximum discrepancy in performance is only 8.9 steps per agent in \texttt{den520d}, and our reproduced model performs better in \texttt{empty} and \texttt{warehouse}, confirming the reliability of our reproduction. For SVM, delay-versus-iteration statistics are not provided, so we compare the ``average ranking". Average ranking defined by \citet{huang2022anytime} is the average rank of the predicted top neighborhood out of 20 candidates. The results, shown in the left half of Table~\ref{tab:validation-svm-nns}, reveal that the discrepancy in average rank is no greater than one, further validating our reproduction. \textit{2) Generalization ability}. It is costly and impossible to train separate models for different parameters, such as the number of agents or neighborhood size. Instead, we wish to train models under one parameter setting and evaluate their performance by generalizing to unseen configurations, a goal also emphasized in their original papers. To assess generalization, we conduct experiments comparing the performance of two models evaluated under the same setting but trained with one key parameter altered. The key parameters include neighborhood size, selection strategy, number of agents, and initial solver. The comparison results are summarized in Table~\ref{table:nns-svm-gen-nb-agent-sol-strategy}. We observe that the final delays produced by the generalized models (trained and tested under different configurations) are generally comparable to those of the ungeneralized models (trained and tested under the same configuration).


\begin{table}[t]
\centering
\vspace{6pt}
\caption{Generalization results on four key parameters of SVM and NNS. The time limit is $300$s.}
\vspace{-6pt}
\setlength{\tabcolsep}{3pt}
\resizebox{0.465\textwidth}{!}{
\label{table:nns-svm-gen-nb-agent-sol-strategy}
\begin{tabular}{@{}c|c|c|rr|rrcc|c|c|rr|rr@{}}
\toprule
\multicolumn{7}{c}{\textbf{Neighborhood Size Generalization}} & \multicolumn{1}{c}{\ } \multirow{7}{*}{\rule{0.8pt}{22.8\normalbaselineskip}}  & \multicolumn{7}{c}{\textbf{Selection Strategy Generalization}} \\ \midrule
 & \multicolumn{2}{c|}{NB} & \multicolumn{2}{c|}{SVM} & \multicolumn{2}{c}{NNS} & \multicolumn{1}{c}{\ }& & \multicolumn{2}{c|}{Removal Strategy} & \multicolumn{2}{c|}{SVM} & \multicolumn{2}{c}{NNS} \\ \midrule
Map & Train & Test & Delay & AUC & Delay & AUC & \multicolumn{1}{c}{\ } & Map & Train & Test & Delay & AUC & Delay & AUC \\ \midrule
random & 8 & \multirow{2}{*}{16} & 1688.2 & 61.3 & 1829.5 & 61.9 & \multicolumn{1}{c}{\ } & random & RW & \multirow{2}{*}{ADP} & 1576.1 & 56.6 & 1752.7 & 58.3 \\
+250 & 16 & & 1700.6 & 60.7 & 1912.3 & 64.8 & \multicolumn{1}{c}{\ } & +250 & ADP & & 1566.2 & 55.3 & 1735.0 & 58.0 \\ \midrule
den520d & 16 & \multirow{2}{*}{8} & 837.6 & 97.8 & 1485.7 & 76.8 & \multicolumn{1}{c}{\ } & den520d & RWP & \multirow{2}{*}{ADP} & 1793.0 & 168.3 & 2759.8 & 126.9 \\
+700 & 8 & & 861.5 & 98.5 & 1150.0 & 62.4 & \multicolumn{1}{c}{\ } & +700 & ADP & & 1662.9 & 164.3 & 1964.9 & 93.2 \\ \toprule
\multicolumn{7}{c}{\textbf{Number of Agents Generalization}} & \multicolumn{1}{c}{\ }  & \multicolumn{7}{c}{\textbf{Initial Solver Generalization}} \\ \midrule
 & \multicolumn{2}{c|}{Agent Num} & \multicolumn{2}{c|}{SVM} & \multicolumn{2}{c}{NNS} &\multicolumn{1}{c}{\ } & & \multicolumn{2}{c|}{Initial Solution} & \multicolumn{2}{c|}{SVM} & \multicolumn{2}{c}{NNS} \\ \midrule
Map & Train & Test & Delay & AUC & Delay & AUC & \multicolumn{1}{c}{\ } & Map & Train & Test & Delay & AUC & Delay & AUC \\ \midrule
\multirow{2}{*}{random} & 250 & \multirow{2}{*}{350} & 4800.2 & 175.8 & 5466.7 & 193.5 & \multicolumn{1}{c}{\ } & random & LNS2 & \multirow{2}{*}{LaCAM2} & 1521.2 & 60.1 & 1747.1 & 63.6 \\
 & 350 & & 4806.8 & 177.7 & 5513.1 & 188.9 & \multicolumn{1}{c}{\ } & +250 & LaCAM2 & & 1517.5 & 61.4 & 1811.4 & 65.7 \\ \midrule
\multirow{2}{*}{random} & 250 & \multirow{2}{*}{250} & 1713.9 & 59.1 & 1683.7 & 57.5 & \multicolumn{1}{c}{\ } & random & LNS2 & \multirow{2}{*}{EECBS} & 1443.6 & 50.0 & 1604.9 & 53.9 \\
 & 350 & & 1578.6 & 56.3 & 1738.5 & 60.4 & \multicolumn{1}{c}{\ } & +250 & EECBS & & 1452.2 & 51.4 & 1591.6 & 53.7 \\ \midrule
\multirow{2}{*}{den520d} & 700 & \multirow{2}{*}{900} & 1816.6 & 216.5 & 1821.4 & 168.1 & \multicolumn{1}{c}{\ } & den520d & LNS2 & \multirow{2}{*}{LaCAM2} & 660.7 & 138.1 & 816.6 & 85.9 \\
 & 900 & & 4171.4 & 329.7 & 1426.5 & 87.6 & \multicolumn{1}{c}{\ } & +700 & LaCAM2 & & 710.9 & 128.0 & 771.5 & 79.1 \\ \midrule
\multirow{2}{*}{den520d} & 900 & \multirow{2}{*}{700} & 931.5 & 134.9 & 793.6 & 67.3 & \multicolumn{1}{c}{\ } & Paris & LNS2 & \multirow{2}{*}{EECBS} & 113.4 & 3.6 & 125.2 & 4.1 \\
 & 700 & & 665.2 & 77.9 & 814.0 & 68.2 & \multicolumn{1}{c}{\ } & +450 & EECBS & & 115.7 & 3.6 & 131.9 & 4.2 \\ \bottomrule
\end{tabular}
}

\vspace{1pt}
\parbox{0.465\textwidth}{\scriptsize Note: `NB' is the neighborhood size. RW, ADP, and RWP stand for RandomWalk, Adaptive, and RandomWalkProb. Every two rows in a block represent a set of generalization experiments on one map.}
\vspace{-10pt}
\end{table}

\section{Findings and Future Directions}
We perform evaluation for aforementioned approaches under the unified framework outlined in Sec.~\ref{sec:unify-setting}, which reveals six key insights that challenge existing research perspectives and highlight four promising directions for future work.

The results are mainly presented as running curves and value tables. For simplicity and clarity, we present representative results when the initial solver is LNS2, with a focus on the highest number of agents in each map, since those are the most congested and challenging cases. We also present cases with a medium number of agents, which are used to train SVM and NNS. Including the case of medium number makes it easier to access the performance of SVM and NNS. The results in other settings reflect similar observations and are deferred to Sec.~\ref{sec:full-results} in Appendix. We also include the results of the reproduced SVM and NNS when they are trained according to their original papers. We add a prefix `Ori-' (e.g., Ori-SVM, Ori-NNS) to distinguish with SVM and NNS trained under our unified setting. 

For rule-based methods, the results are obtained using the best-performing neighborhood size. For evaluating SVM and NNS, we select the optimal combination of heuristic and neighborhood size for proposing neighborhood candidates. For example, in \texttt{empty} map with 500 agents, RandomWalk with neighborhood size 8 achieves the lowest delay within $300$s. Consequently, we use RandomWalk to generate neighborhood candidates for SVM and NNS, and employ neighborhood size 8 during execution in that scenario. For evaluating Ori-SVM, and Ori-NNS, we adhere to the heuristic and neighborhood size specified in their original papers.

\subsection{Key Insights and Analyses} \label{sec-result-finding}

\begin{figure}[t!]
\hspace*{-4pt} 
\raggedleft
\includegraphics[width=0.484\textwidth]{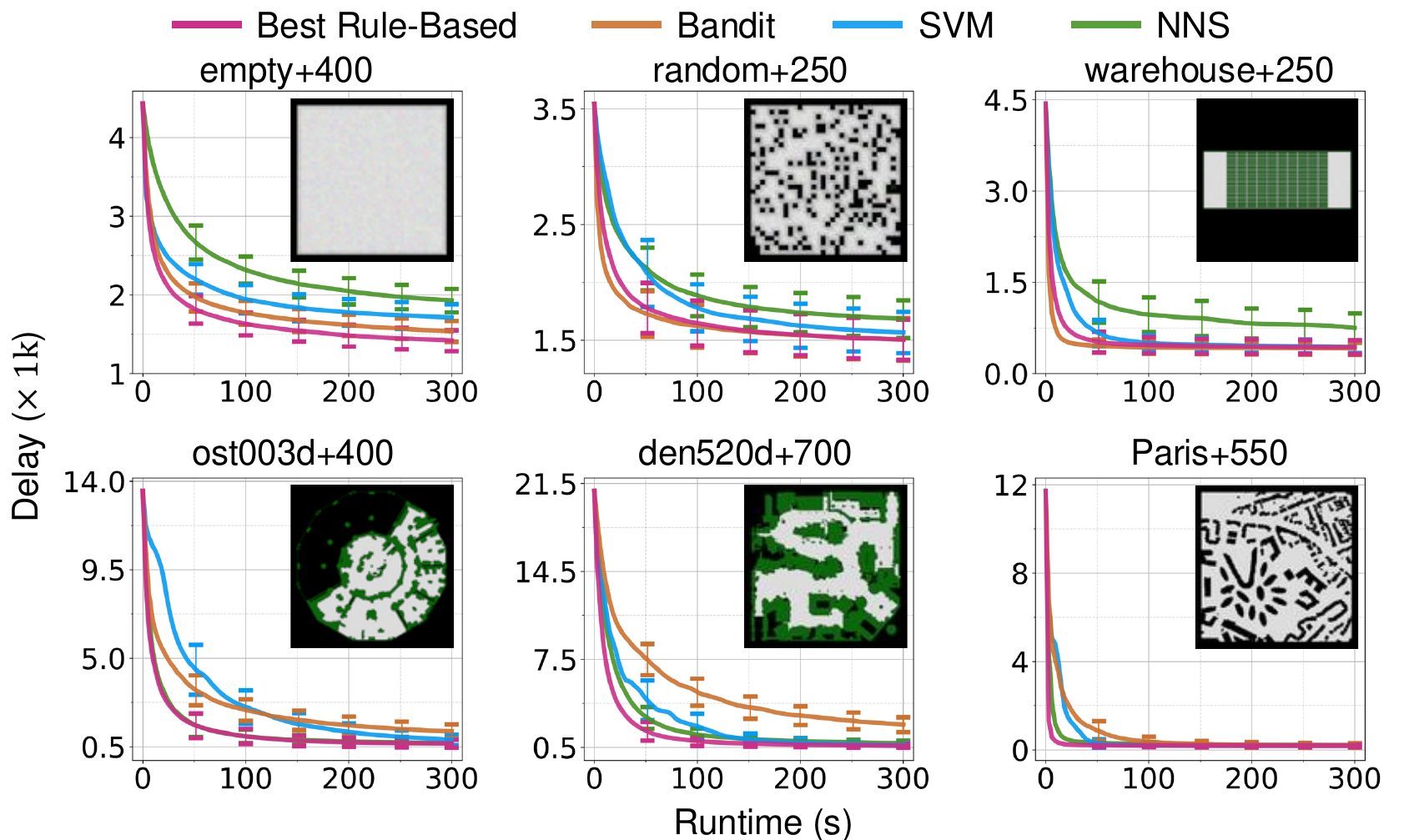}
\vspace{-16pt}
\caption{\textbf{Delay vs. Time in maps with a Medium number of agents}. Error bars represent the variance of delay across $25$ different scenes. The best rule-based strategies are Adaptive for \texttt{empty} and \texttt{warehouse}, RandomWalk for \texttt{random}, and RandomWalkProb for \texttt{ost003d}, \texttt{den520d}, and \texttt{Paris}.}
\label{fig:delay-time}
\vspace{-10pt}
\end{figure}

\begin{table*}[t!]
    \caption{Final delays and AUC across methods in maps with the highest and medium number of agents. Time limits are $300$s and $60$s, respectively. Initial solver is LNS2.}
    \vspace{-8pt}
    \label{table:combined-agent-lns2-delay-auc}
    \centering
    \setlength{\tabcolsep}{2pt} 
    \resizebox{\textwidth}{!}{%
    
    \begin{tabular}{l|ccccccrrrrrr|rrrrrrrrrrrr}
    \toprule
     & \multicolumn{12}{c|}{\textbf{ Highest Number of Agents; Time: 300s}} & \multicolumn{12}{c}{\textbf{Medium Number of Agents; Time: 300s}} \label{row:mid-time300s} \\
    \midrule
     \multirow{2}{*}{\textbf{Methods}} & \multicolumn{2}{c}{\textbf{empty}+500} & \multicolumn{2}{c}{\textbf{random}+350} & \multicolumn{2}{c}{\textbf{warehouse}+350} & \multicolumn{2}{c}{\textbf{ost003d}+600} & \multicolumn{2}{c}{\textbf{den520d}+900} & \multicolumn{2}{c|}{\textbf{Paris}+750} 
     & \multicolumn{2}{c}{\textbf{empty}+400} & \multicolumn{2}{c}{\textbf{random}+250} & \multicolumn{2}{c}{\textbf{warehouse}+250} & \multicolumn{2}{c}{\textbf{ost003d}+400} & \multicolumn{2}{c}{\textbf{den520d}+700} & \multicolumn{2}{c}{\textbf{Paris}+550} \\
     \cmidrule(r){2-13} \cmidrule(r){14-25}
     
     & Delay & AUC & Delay & AUC & Delay & AUC & Delay & AUC & Delay & AUC & Delay & AUC 
     & Delay & AUC & Delay & AUC & Delay & AUC \makebox[5pt][l] & Delay & AUC & Delay & AUC & Delay & AUC\\
    \midrule
    RW & \textcolor{Red}{4050.5} & 145.4 & 4439.2 & \textcolor{cyan}{150.4} & \textcolor{Red}{1041.8} & \textcolor{cyan}{41.3} & 6069.5 & 318.3 & 2290.1 & 166.7 & 404.7 & \textcolor{cyan}{25.4} 
    & \textcolor{Red}{1397.2} & \textcolor{Red}{49.1} & \textcolor{Red}{1504.5} & \textcolor{cyan}{50.4} & 433.1 & 15.3\makebox[8pt][l] & 761.8 & 48.3 & 934.1 & 80.3 & 203.6 & \textcolor{cyan}{9.1}\\
    INT & 4205.4 & 143.9 & 4609.3 & 155.5 & 1949.2 & 84.6 & 8824.9 & 404.8 & 6776.0 & 347.7 & 1680.8 & 110.9 
    & 1513.7 & 53.7 & 1582.6 & 53.2 & 695.2 & 32.8\makebox[8pt][l] & 2048.3 & 100.6 & 3415.0 & 175.9 & 540.8 & 38.3\\
    RAND & 4438.9 & 155.5 & 4635.2 & 167.8 & 1515.9 & 66.2 & 8424.3 & 399.2 & 6447.9 & 345.6 & 1211.7 & 99.1 
    & 1703.2 & 59.6 & 1606.3 & 53.3 & 537.7 & 23.4\makebox[8pt][l] & 1668.8 & 86.3 & 2907.4 & 171.7 & 345.4 & 33.7\\
    ADP & 4093.8 & \textcolor{Red}{140.4} & \textcolor{cyan}{4432.8} & \textcolor{Red}{150.1} & 1073.0 & 44.1 & 6325.9 & 323.0 & 2611.1 & 187.9 & 396.9 & 33.7 
    & \textcolor{cyan}{1418.6} & \textcolor{cyan}{50.0} & 1527.5 & 50.6 & 435.6 & 16.2\makebox[8pt][l] & 746.1 & 51.0 & 1049.8 & 88.0 & 192.2 & 11.7\\
    RWP & \textcolor{cyan}{4051.3} & \textcolor{cyan}{143.9} &  \textcolor{Red}{4408.2} & 156.6 & 1134.9 & 44.7 & \textcolor{Red}{4731.3} & \textcolor{Red}{260.7} & \textcolor{Red}{1387.2} & \textcolor{Red}{130.6} & \textcolor{Red}{375.6} & \textcolor{Red}{21.8} 
    & 1431.1 & 50.1 & 1534.4 & 50.9 & 443.8 & 16.4\makebox[8pt][l] & \textcolor{Red}{652.3} & \textcolor{Red}{39.3} & \textcolor{Red}{620.2} & \textcolor{Red}{48.3} & \textcolor{Red}{183.8} & \textcolor{Red}{7.4}\\
    \midrule
    SVM & 5053.2 & 167.1 & 4800.2 & 175.8 & 1107.4 & 65.7 & 10104.8 & 459.1 & \textcolor{cyan}{1816.6} & 216.5 & \textcolor{cyan}{388.7} & 65.3 
    & 1588.3 & 64.5 & 1713.9 & 59.1 & 439.6 & 19.3\makebox[8pt][l] & 850.3 & 84.3 & \textcolor{cyan}{665.2} & 77.9 & \textcolor{cyan}{184.7} & 16.5\\
    NNS & 4803.2 & 179.7 & 5466.7 & 193.5 & 1871.1 & 77.1 & \textcolor{cyan}{5501.2} & 294.6 & 1821.4 & \textcolor{cyan}{168.1} & 468.2 & 36.6 
    & 1928.5 & 70.0 & 1683.7 & 57.5 & 749.0 & 31.4\makebox[8pt][l] & \textcolor{cyan}{679.5} & \textcolor{cyan}{41.2} & 814.0 & \textcolor{cyan}{68.2} & 217.5  & 12.2\\
    Bandit & 4318.5 & 150.8 & 4564.1 & 155.8 & \textcolor{cyan}{1047.6} & \textcolor{Red}{38.1} & 6093.6 & \textcolor{cyan}{311.4} & 5343.4 & 308.8 & 576.9 & 70.6 
    & 1537.2 & 54.2 & \textcolor{cyan}{1507.3} & \textcolor{Red}{49.3} & \textcolor{Red}{414.2} & \textcolor{Red}{14.1}\makebox[8pt][l] & 1276.3 & 75.2 & 2297.3 & 150.2 & 205.3 & 20.7\\
    Ori-SVM & 4776.3 & 174.7 & 5105.2 & 186.3 & 1100.9 & 64.6 & 8604.1 & 457.2 & 6161.2 & 420.2 & 483.2 & 126.6 
    & 1640.4 & 63.9 & 1662.8 & 58.3 & 441.0 & 21.1\makebox[8pt][l] & 1086.3 & 106.5 & 1880.1 & 200.1 & 196.5 & 37.6\\
    Ori-NNS & 5305.6 & 186.4 & 6004.1 & 206.3 & 1094.5 & 56.1 & 11426.4 & 517.9 & 13367.1 & 624.8 & 2409.4 & 243.5 
    & 1464.8 & 56.8 & 1528.0 & 54.8 & \textcolor{cyan}{417.4} & 17.7\makebox[8pt][l] & 1483.2 & 109.0 & 4228.7 & 286.7 & 721.5 & 67.7\\
    \midrule
     & \multicolumn{12}{c|}{\textbf{Highest Number of Agents; Time: 60s}} & \multicolumn{12}{c}{\textbf{Medium Number of Agents; Time: 60s}} \\
     
    \midrule
    RW & \textcolor{Red}{4949.4} & \textcolor{Red}{34.9} & \textcolor{cyan}{5280.3} & 37.7 & \textcolor{cyan}{1331.4} & \textcolor{cyan}{14.1} & 13845.4 & 110.0 & 7159.6 & \textcolor{cyan}{84.2} & \textcolor{cyan}{559.5} & \textcolor{cyan}{14.5} 
    & \textcolor{Red}{1726.2} & \textcolor{Red}{13.2} & \textcolor{cyan}{1744.9} & 12.5 & \textcolor{cyan}{472.7} & \textcolor{cyan}{4.5}\makebox[8pt][l] & 1785.4 & \textcolor{cyan}{24.3} & 2583.3 & \textcolor{cyan}{38.4} & \textcolor{cyan}{216.2} & \textcolor{cyan}{4.1}\\
    INT & 5117.5 & 35.9 & 5456.2 & 38.4 & 3346.2 & 26.9 & 16830.7 & 127.1 & 14664.4 & 127.0 & 4604.3 & 49.8 
    & 1900.5 & 14.3 & 1862.1 & 13.2 & 1333.8 & 11.7\makebox[8pt][l] & 3975.2 & 38.2 & 7100.6 & 68.3 & 1494.2 & 19.8\\
    RAND & 5376.6 & 36.2 & 5416.5 & 37.4 & 2420.4 & 23.5 & 16765.7 & 127.9 & 14546.4 & 124.7 & 4620.3 & 45.2 
    & 2105.9 & 15.2 & 1841.5 & 13.1 & 791.7 & 9.1\makebox[8pt][l] & 3482.2 & 34.5 & 7256.5 & 67.9 & 1485.0 & 18.7\\
    ADP & \textcolor{cyan}{4997.9} & \textcolor{cyan}{35.0} & 5290.9 & \textcolor{cyan}{37.2} & 1463.4 & 15.4 & 14271.5 & 108.2 & 8506.6 & 89.3 & 689.1 & 16.8 
    & 1768.8 & 13.3 & 1762.5 & \textcolor{cyan}{12.3} & 488.3 & 5.2\makebox[8pt][l] & 2013.8 & 25.3 & 3317.7 & 43.1 & 236.4 & 5.2\\
    RWP & 5024.8 & 35.5 & \textcolor{Red}{5224.1} & \textcolor{Red}{37.1} & 1455.6 & 14.3 & \textcolor{Red}{11123.2} & \textcolor{Red}{99.2} & \textcolor{Red}{4547.5} & \textcolor{Red}{67.6} & \textcolor{Red}{549.0} & \textcolor{Red}{10.7} 
    & \textcolor{cyan}{1765.6} & \textcolor{cyan}{13.4} & 1761.4 & 12.5 & 534.6 & 5.1\makebox[8pt][l] & \textcolor{Red}{1127.0} & \textcolor{Red}{16.9} & \textcolor{Red}{1498.7} & \textcolor{Red}{28.9} & \textcolor{Red}{200.6} & \textcolor{Red}{2.9}\\
    \midrule
    SVM & 6565.9 & 43.7 & 6543.7 & 46.4 & 3105.0 & 31.6 & 18765.4 & 130.6 & 10998.2 & 111.5 & 2267.7 & 50.0 
    & 2584.0 & 19.7 & 1990.0 & 14.8 & 600.5 & 8.0\makebox[8pt][l] & 4096.8 & 44.1 & 3562.8 & 46.6 & 243.5 & 11.8\\
    NNS & 6355.9 & 43.2 & 6877.1 & 46.4 & 2677.8 & 22.7 & \textcolor{cyan}{12267.2} & \textcolor{cyan}{102.9} & \textcolor{cyan}{6444.6} & 85.6 & 907.1 & 20.6 
    & 2461.4 & 18.1 & 2049.3 & 14.6 & 1018.5 & 9.4\makebox[8pt][l] & \textcolor{cyan}{1338.2} & \textcolor{cyan}{19.5} & \textcolor{cyan}{2115.6} & 38.8 & 292.2 & 5.8\\
    Bandit & 5473.0 & 38.0 & 5540.1 & 38.5 & \textcolor{Red}{1216.4} & \textcolor{Red}{12.0} & 13217.6 & 106.2 & 13552.1 & 115.6 & 3220.9 & 41.1 
    & 1916.4 & 14.3 & \textcolor{Red}{1699.9} & \textcolor{Red}{11.7} & \textcolor{Red}{446.8} & \textcolor{Red}{4.0}\makebox[8pt][l]& 3127.5 & 31.4 & 6821.9 & 63.3 & 692.2 & 13.9\\
    Ori-SVM & 6566.5 & 44.6 & 7113.9 & 47.7 & 2971.2 & 30.5 & 20754.1 & 140.7 & 20674.6 & 153.4 & 7769.9 & 82.1 
    & 2416.5 & 18.9 & 2176.5 & 14.8 & 657.2 & 10.0\makebox[8pt][l] & 5713.4 & 53.8 & 11421.6 & 93.5 & 1628.3 & 29.6\\
    Ori-NNS & 6908.0 & 46.0 & 7514.5 & 48.9 & 2182.7 & 24.1 & 21677.9 & 144.8 & 26155.1 & 172.3 & 12831.9 & 97.3 & 2127.1 & 17.3 & 1973.9 & 14.6 & 531.7 & 7.0\makebox[8pt][l] & 5246.7 & 49.7 & 13704.0 & 102.6 & 3342.0 & 39.5\\
    
    \bottomrule
    \end{tabular}

    }
    
    \vspace{1pt}
    
    \parbox{\textwidth}{\scriptsize Note: RW, INT, RAND, ADP, and RWP stand for RandomWalk, Intersection, Random, Adaptive, and RandomWalkProb, respectively. The number of agents follows the name of a map, i.e., after ``+". Highlighted are the results ranked the \textcolor{Red}{first}, and \textcolor{cyan}{second}.}
    \vspace{-6pt}
    \end{table*}
    

\textbf{1) Rule-based strategies are strong competitors to learning-based strategies in terms of time efficiency}. Note that the final delays we report for rule-based strategies are generally lower than those in previous studies, as we only measure the time spent on core processes. This same time measurement scheme is applied to learning-based strategies, ensuring a fair comparison.  Table~\ref{table:combined-agent-lns2-delay-auc} presents the final delays and AUCs of delay-versus-time curves when the time limits are $300$s and $60$s (with the highest number of agents in each domain). Rule-based strategies achieve the best final delays in $83.3\%$ (20/24) of the cases. To minimize the impact of generalizing to untrained scenarios on the performance of SVM and NNS, we also investigate their performance on maps with a medium number of agents where training data are collected. The delay-versus-time curves for these scenarios are shown in Figure~\ref{fig:delay-time}. For clarity, only the curves of the best-performing rule-based strategies are included. NNS matches the efficiency of the best rule-based strategy only in the \texttt{ost003d} map. In other domains, SVM and NNS are generally slower.

Our findings contradict those of learning-based studies. For example, \citet{huang2022anytime} claimed that SVM-based neighborhood prediction outperforms rule-based methods in terms of time efficiency. However, under our unified evaluation framework, SVM shows no clear advantage over rule-based ones. Additionally, \citet{phan2024adaptive} directly adopted the results from~\citet{huang2022anytime} and suggested that Bandit outperforms SVM. In our evaluation, however, Bandit is better than SVM in some cases but worse in others. We also notice that the final delays for rule-based methods reported in \citet{huang2022anytime} are significantly higher than those in \citet{li2021anytime}, and their time measurement schemes are not clearly described. This raises the possibility that rule-based strategies were under-reported or evaluated using inconsistent time measurement schemes.  \citet{yan2024neural} focused solely on cases where PBS outperforms PP and aimed to improve these cases using deep learning (though our evaluation shows PP is generally better than PBS). Consequently, their application scope is limited. When applied to diverse scenarios in our evaluation, NNS does not demonstrate faster performance than rule-based approaches.

\noindent\textbf{2) SVM and NNS incur high overheads compared to PP replanning}. To understand the time inefficiency of SVM and NNS, we analyze their additional overheads compared to rule-based methods. In each iteration, SVM and NNS introduce two main additional sources of overheads: \textit{1) Proposition}, which involves generating a set of neighborhood candidates using rule-based strategies ($20$ candidates in SVM, and $100$ in NNS); \textit{2) Prediction}, which predicts the best neighborhood using trained ranking models. Table~\ref{table:max-agent-lns2-overheads} summarizes the percentage of these overheads in total time used for maps with the highest number of agents. 

The proposition overhead in SVM is negligible due to the small number of candidates (i.e., $20$), but it increases greatly in NNS due to a larger candidate pool (i.e., $100$). The prediction overhead is notably high for both SVM and NNS. This is because PP replanning is super fast per iteration, making the prediction speed of the machine learning models the bottleneck.  We also report the overheads of Ori-NNS when PBS is used for replanning. In this case, the proposition and prediction overheads in NNS become relatively small because PBS takes a longer time to execute. This observation supports \citet{yan2024neural}'s approach of applying deep learning only in cases where PBS performs better, as the neural network overhead becomes acceptable with PBS replanning. However, this further highlights the limited application scope of NNS.






\begin{table}[h]
\caption{Percentage of proposition and prediction overheads in SVM and NNS. Initial solver is LNS2.}
\vspace{-8pt}
\label{table:max-agent-lns2-overheads}
\centering
\setlength{\tabcolsep}{1.5pt}
\resizebox{0.465\textwidth}{!}{%
\begin{tabular}{l|rrrrrrrrrrrr}
\toprule
 \multirow{2}{*}{\textbf{Methods}} & \multicolumn{2}{c}{\textbf{empty}+500} & \multicolumn{2}{c}{\textbf{random}+350} & \multicolumn{2}{c}{\textbf{warehouse}+350} & \multicolumn{2}{c}{\textbf{ost003d}+600} & \multicolumn{2}{c}{\textbf{den520d}+900} & \multicolumn{2}{c}{\textbf{Paris}+750} \\
 \cmidrule(r){2-3} \cmidrule(r){4-5} \cmidrule(r){6-7} \cmidrule(r){8-9} \cmidrule(r){10-11} \cmidrule(r){12-13}
 & Prop & Pred & Prop & Pred & \makebox[15pt][r]Prop & Pred\makebox[10pt][l] & Prop & Pred & Prop & Pred & Prop & Pred\\
\midrule
SVM & $0.4\%$ & $41.4\%$ & $3.2\%$ & $38.2\%$ & $0.4\%$ & $33.4\%$ \makebox[3pt][l]& $1.9\%$ & $33.5\%$ & $1.7\%$ & $45.0\%$ & $1.5\%$ & $35.1\%$\\
NNS & $7.5\%$ & $53.2\%$  & $11.3\%$ & $46.8\%$  & $16.7\%$ & $47.6\%$ \makebox[3pt][l]& $7.6\%$ & $13.1\%$ & $16.3\%$ & $27.8\%$ & $34.8\%$ & $29.3\%$\\
Ori-NNS & $8.4\%$ & $1.4\%$ & $4.7\%$ & $3.4\%$ & $0.8\%$ & $0.6\%$ \makebox[3pt][l]& $0.8\%$ & $1.9\%$ & $0.4\%$ & $0.7\%$ & $2.3\%$ & $1.4\%$\\

\bottomrule
\end{tabular}
}

\parbox{0.465\textwidth}{\scriptsize Note: ’Prop’ represents proposition. ’Pred’ represents prediction.}


\end{table}

\begin{figure}[t]

\centering
\includegraphics[width=0.48\textwidth]{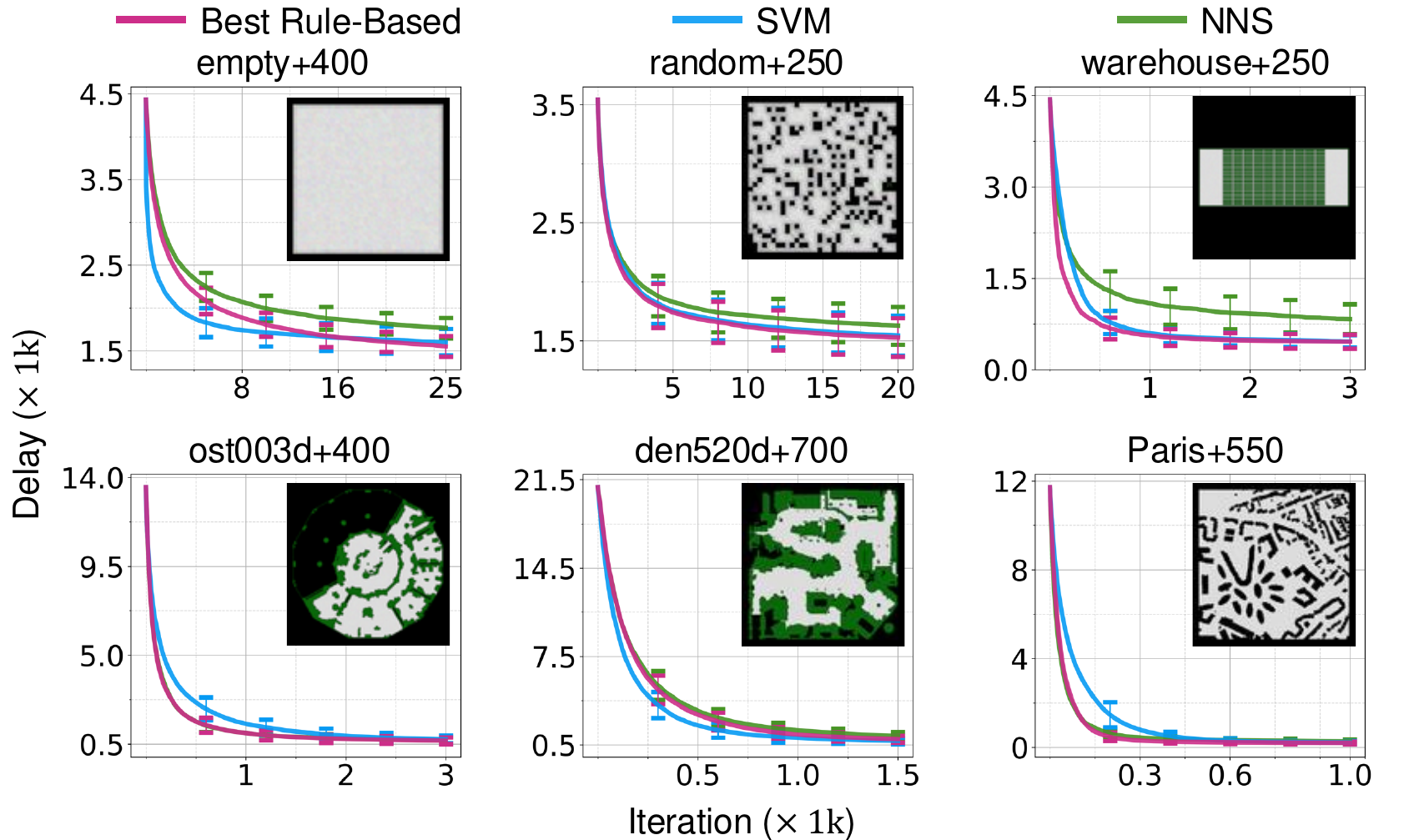}
\vspace{-19pt}
\caption{\textbf{Delay vs. Iteration in maps with a Medium number of agents}. Error bars represent the variance of delay across $25$ different scenes. The best rule-based strategies are Adaptive for \texttt{empty} and \texttt{warehouse}, RandomWalk for \texttt{random}, and RandomWalkProb for \texttt{ost003d}, \texttt{den520d}, and \texttt{Paris}.}
\label{fig:delay-iter-svm-nns}
\vspace{-10pt}
\end{figure}

\noindent\textbf{3) The improvement capacity of supervised learning methods per iteration is limited}. As discussed above, SVM and NNS introduce high time overheads. Here, we investigate their improvement capacity per iteration, which is independent of the time overheads. We compare them with the best rule-based strategies using the \textit{delay-versus-iteration} criterion in maps with medium number of agents. Note that the best rule-based strategies in these scenarios are employed to generate training data for SVM and NNS. The performance curves are shown in Figure~\ref{fig:delay-iter-svm-nns}.

SVM and NNS aim to predict and select the best neighborhood from the candidate pool for replanning in each iteration. However, as illustrated in Figure~\ref {fig:delay-iter-svm-nns}, only in \texttt{empty} and \texttt{den520d}, SVM is able to predict a better neighborhood than the one picked by rule-based methods. In other cases, the improvement capacity of supervised learning models fails to surpass that of rule-based strategies.  Also, as indicated in Table~\ref{tab:validation-svm-nns}, the trained ranking models struggle to accurately select the ground truth best neighborhood. This suggests that achieving a clear advantage over rule-based strategies requires a more powerful neighborhood ranking model with higher prediction accuracy.

\noindent\textbf{4) Under-performance of Bandit vs. Adaptive}. Bandit is another learning-based method, but it employs reinforcement learning instead of supervised learning. As a result, it requires no training data and dynamically identifies the best rule-based strategies in each iteration during execution. This allows us to view it as another variant of Adaptive, where bandit logits serve as weights for selecting rule-based strategies. 

When comparing Bandit and Adaptive in Table~\ref{table:combined-agent-lns2-delay-auc}, Bandit performs worse than Adaptive in $66.6\%$ cases (16/24) in terms of final delays. The performance gap between them is particularly significant in map \texttt{den520d}. As discussed in Sec.~\ref{sec:discussion-learning}, Bandit uses a non-contextual algorithm, whereas a contextual algorithm is more appropriate, which may explain its under-performance. However, we also observe that with a short time limit, i.e., $60$s, Bandit achieves the lowest delay and AUC in maps like \texttt{random} and \texttt{warehouse}. This suggests that properly switching among rule-based strategies and neighborhood sizes can accelerate MAPF-LNS. 

\noindent\textbf{5) RandomWalk (with its variant RandomWalkProb) shows robust performance across diverse scenarios}. The complete results for final delays across various maps and number of agents are provided in Sec.~\ref{sec:full-results} in Appendix. In the majority of cases, RandomWalk (with its variant RandomWalkProb) achieves the best final delays. Even in scenarios where it does not rank first, its performance remains close to the best. This observation slightly contrasts with the findings of \citet{li2021anytime}, where Adaptive outperformed RandomWalk in more than half of the cases. While the time measurement scheme in \citet{li2021anytime} is not clearly specified, under our unified evaluation framework, RandomWalk and its variant exhibit a clear advantage over other methods.

\noindent\textbf{6) Quality of the initial solution is not highly critical}. In our experiments, we study two initial solvers: LNS2 and LaCAM2. LaCAM2 is faster but generally produces low-quality solutions compared to LNS2. A comparison of their delays in maps with the highest number of agents is provided in Table~\ref{table:max-agent-lacam2-lns2-init-delay}. However, running various LNS methods starting from LaCAM2 initial solutions yields final delays similar to those starting from LNS2 initial solutions. The results using LaCAM2 as the initial solver are shown in Table~\ref{table:lacam2-as-initial}. 

Comparing Table~\ref{table:combined-agent-lns2-delay-auc} (LNS2) and Table~\ref{table:lacam2-as-initial} (LaCAM2), the discrepancies in final delays for RandomWalkProb within $300$s are consistently less than 350 across all maps. Even under a $60$s limit, the discrepancies remain below 770, except for map \texttt{ost003d}. We also examine the final delays when using EECBS as the initial solver, as shown in Table~\ref{table:eecbs-as-initial} in Appendix. EECBS provides better solution quality than LNS2 but is unable to solve all cases in $10$s. Comparing Table~\ref{table:combined-agent-lns2-delay-auc} (LNS2) and \ref{table:eecbs-as-initial} (EECBS), the differences in final delays for RandomWalkProb within both $300$s and $60$s are consistently below 400. These findings suggest that delays decrease rapidly in the early stages of the LNS process. Consequently, fast and scalable solvers like LaCAM2 and LNS2, which can solve most MAPF instances with many agents, should be preferred even if their initial solutions are of lower quality.


\begin{table}[t!]
\caption{Delays of the initial solutions found by LNS2 and LaCAM2 in maps with the highest number of agent.}

\vspace{-6pt}

\label{table:max-agent-lacam2-lns2-init-delay}
\centering
\setlength{\tabcolsep}{2pt}
\resizebox{0.45\textwidth}{!}{%
\begin{tabular}{l|rrrrrr}
\toprule
 \textbf{Solver} & \makecell{\textbf{empty} \\ \textbf{+500}} & \makecell{\textbf{random} \\ \textbf{+350}} & \makecell{\textbf{warehouse} \\ \textbf{+350}} & \makecell{\textbf{ost003d} \\ \textbf{+600}} & \makecell{\textbf{den520d} \\ \textbf{+900}} & \makecell{\textbf{Paris} \\ \textbf{+750}} \\
\midrule
LNS2 & 8724.2 & 9305.4 & 8020.1 \makebox[5pt][l]& 26806.3 & 31463.2 & 20460.5 \\
LaCAM2 & 13058.5 & 14969.3 & 22804.4 \makebox[5pt][l]& 38632.1 & 51204.5 & 31249.1 \\
\bottomrule
\end{tabular}
}
\vspace{-6pt}
\end{table}

\begin{table}[t!]
\caption{Final delays across methods in maps with the highest and medium number of agents. Time limits are $300$s and $60$s, respectively. Initial solver is LaCAM2.}
\vspace{-8pt}
\label{table:lacam2-as-initial}
\centering
\setlength{\tabcolsep}{2pt}
\resizebox{0.46\textwidth}{!}{
\begin{tabular}{l|rrrrrr|rrrrrr}
\toprule
\multicolumn{1}{c}{}  & \multicolumn{6}{|c|}{\textbf{Largest Number of Agents; Time: 300s}} & \multicolumn{6}{c}{\textbf{Medium Number of Agents; Time: 300s}}\\
\midrule
\multirow{1}{*}{\textbf{Methods}}  
& \makecell{\textbf{empty} \\ \textbf{+500}} & \makecell{\textbf{random} \\ \textbf{+350}} & \makecell{\textbf{warehouse} \\ \textbf{+350}} & \makecell{\textbf{ost003d} \\ \textbf{+600}} & \makecell{\textbf{den520d} \\ \textbf{+900}} & \makecell{\textbf{Paris} \\ \textbf{+750}} 
& \makecell{\textbf{empty} \\ \textbf{+400}} & \makecell{\textbf{random} \\ \textbf{+250}} & \makecell{\textbf{warehouse} \\ \textbf{+250}} & \makecell{\textbf{ost003d} \\ \textbf{+400}} & \makecell{\textbf{den520d} \\ \textbf{+700}} & \makecell{\textbf{Paris} \\ \textbf{+550}} \\
\midrule
RW &  \textcolor{Red}{4237.7} & \textcolor{Red}{4341.6} & \textcolor{cyan}{1043.0}\makebox[5pt][l]{} & 7514.0 & 2486.1 & 373.2 & \textcolor{Red}{1447.9} & \textcolor{cyan}{1472.7} & 430.0\makebox[5pt][l]{} & 430.0 & 933.1 & 198.0 \\
INT & 4448.5 & 4622.2 & 2011.0\makebox[5pt][l]{} & 10394.8 & 6426.0 & 1821.2 & 1593.4 & 1537.2 & 698.5\makebox[5pt][l]{} & 1849.6 & 3163.0 & 776.5 \\
RAND & 4828.2 & 4683.9 & 1509.0\makebox[5pt][l]{} & 12466.1 & 6923.5 & 1362.6 & 1756.2 & 1613.5 & 562.4\makebox[5pt][l]{} & 1488.2 & 2782.7 & 466.9 \\
ADP & 4374.5 & 4406.5 & 1055.0\makebox[5pt][l]{} & 8216.9 & 2819.9 & 385.8 & \textcolor{cyan}{1476.5} & 1506.7 & \textcolor{Red}{424.9}\makebox[5pt][l]{} & 805.2 & 1075.2 & 194.8 \\
RWP & \textcolor{cyan}{4335.7} & \textcolor{cyan}{4367.5} & 1130.8\makebox[5pt][l]{} & \textcolor{Red}{5875.0} & \textcolor{Red}{1382.0} & \textcolor{Red}{366.0} & 1484.4 & 1494.1 & 446.7\makebox[5pt][l]{} & \textcolor{Red}{622.9} & \textcolor{Red}{622.4} & \textcolor{Red}{183.3} \\
\midrule
SVM & 5194.9 & 5162.7 & \textcolor{Red}{1042.4}\makebox[5pt][l]{} & 13066.6 & \textcolor{cyan}{2058.6} & \textcolor{cyan}{369.9} & 1692.4 & 1521.2 & 444.6\makebox[5pt][l]{} & 762.5 & \textcolor{cyan}{660.7} & \textcolor{cyan}{184.7} \\
NNS & 5808.0 & 5886.5 & 1907.0\makebox[5pt][l]{} & 8469.1 & 2067.1 & 498.8 & 2094.8 & 1747.1 & 782.1\makebox[5pt][l]{} & \textcolor{cyan}{675.3} & 816.6 & 228.9 \\
Bandit & 4772.6 & 4598.5 & 1067.6\makebox[5pt][l]{} & \textcolor{cyan}{7244.6} & 5595.4 & 627.3 & 1579.0 & \textcolor{Red}{1456.3} & \textcolor{cyan}{426.2}\makebox[5pt][l]{} & 1253.3 & 2439.2 & 210.2 \\
Ori-SVM & 5793.6 & 5983.6 & 1097.6\makebox[5pt][l]{} & 10153.9 & 15199.0 & 616.3 & 1857.5 & 1604.8 & 445.1\makebox[5pt][l]{} & 1083.7 & 2259.0 & 200.6 \\
Ori-NNS & 6085.3 & 7438.8 & 5255.1\makebox[5pt][l]{} & 25804.5 & 31694.0 & 13831.5 & 1614.3 & 1698.5 & 961.5\makebox[5pt][l]{} & 5277.4 & 15529.0 & 2643.7 \\
\midrule
\multicolumn{1}{c}{}  & \multicolumn{6}{|c|}{\textbf{Largest Number of Agents; Time: 60s}} & \multicolumn{6}{c}{\textbf{Medium Number of Agents; Time: 60s}}\\
\midrule
\multirow{1}{*}{\textbf{Methods}}  
& \makecell{\textbf{empty} \\ \textbf{+500}} & \makecell{\textbf{random} \\ \textbf{+350}} & \makecell{\textbf{warehouse} \\ \textbf{+350}} & \makecell{\textbf{ost003d} \\ \textbf{+600}} & \makecell{\textbf{den520d} \\ \textbf{+900}} & \makecell{\textbf{Paris} \\ \textbf{+750}} 
& \makecell{\textbf{empty} \\ \textbf{+400}} & \makecell{\textbf{random} \\ \textbf{+250}} & \makecell{\textbf{warehouse} \\ \textbf{+250}} & \makecell{\textbf{ost003d} \\ \textbf{+400}} & \makecell{\textbf{den520d} \\ \textbf{+700}} & \makecell{\textbf{Paris} \\ \textbf{+550}} \\
\midrule
RW & \textcolor{Red}{5605.0} & \textcolor{cyan}{5535.2} & \textcolor{cyan}{1421.0}\makebox[5pt][l]{} & 20549.4 & \textcolor{cyan}{8274.1} & \textcolor{cyan}{471.4} & \textcolor{Red}{1845.7} & \textcolor{cyan}{1698.3} & \textcolor{cyan}{470.1}\makebox[5pt][l]{} & 1717.2 & \textcolor{cyan}{2642.0} & \textcolor{cyan}{217.2} \\
INT & 5895.3 & 6027.2 & 7091.6\makebox[5pt][l]{} & 21653.0 & 13091.0 & 4794.1 & 2030.1 & 1830.5 & 1731.4\makebox[5pt][l]{} & 3618.8 & 6254.7 & 2269.4 \\
RAND & 6076.0 & 5764.9 & 2764.4\makebox[5pt][l]{} & 22508.1 & 16636.2 & 5078.5 & 2130.5 & 1832.2 & 813.2\makebox[5pt][l]{} & 3750.1 & 7336.5 & 2110.1 \\
ADP & \textcolor{cyan}{5631.0} & 5544.0 & 1577.1\makebox[5pt][l]{} & 20281.3 & 9786.3 & 681.7 & \textcolor{cyan}{1861.8} & 1733.5 & 485.4\makebox[5pt][l]{} & 2023.8 & 3418.0 & 236.4 \\
RWP & 5790.6 & \textcolor{Red}{5463.2} & 1514.4\makebox[5pt][l]{} & \textcolor{cyan}{17560.6} & \textcolor{Red}{4788.8} & \textcolor{Red}{462.8} & 1877.6 & 1711.6 & 537.1\makebox[5pt][l]{} & \textcolor{Red}{1143.8} & \textcolor{Red}{1494.6} & \textcolor{Red}{202.5} \\
\midrule
SVM & 8973.6 & 10463.7 & 4450.8\makebox[5pt][l]{} & 26121.5 & 18143.9 & 5088.3 & 3679.9 & 2274.6 & 1144.1\makebox[5pt][l]{} & 5073.1 & 6070.3 & 545.6 \\
NNS & 8475.4 & 9710.6 & 3366.5\makebox[5pt][l]{} & 20302.9 & 9199.7 & 1080.6 & 2981.6 & 2246.3 & 1201.9\makebox[5pt][l]{} & \textcolor{cyan}{1541.2} & 2934.0 & 350.3 \\
Bandit & 6294.5 & 5944.0 & \textcolor{Red}{1334.3}\makebox[5pt][l]{} & \textcolor{Red}{16993.8} & 14398.9 & 3743.0 & 2027.8 & \textcolor{Red}{1652.8} & \textcolor{Red}{466.8}\makebox[5pt][l]{} & 3118.4 & 7198.2 & 960.8 \\
Ori-SVM & 9866.2 & 10815.6 & 5331.1\makebox[5pt][l]{} & 29866.4 & 30208.9 & 18276.6 & 4503.6 & 2587.1 & 873.0\makebox[5pt][l]{} & 6332.8 & 19781.8 & 7313.9 \\
Ori-NNS & 9953.8 & 12978.2 & 17479.6\makebox[5pt][l]{} & 36312.5 & 47864.2 & 27245.0 & 2942.8 & 2674.5 & 7424.5\makebox[5pt][l]{} & 14720.5 & 30196.8 & 13938.8 \\
\bottomrule
\end{tabular}
}

\parbox{0.46\textwidth}{\scriptsize Note: RW, INT, RAND, ADP, and RWP stand for RandomWalk, Intersection, Random, Adaptive, and RandomWalkProb. The number of agents follows the name of a map, i.e., after ``+". Highlighted are the results ranked the \textcolor{Red}{first}, and \textcolor{cyan}{second}.}
\end{table}

\subsection{Outlooks on Improving MAPF-LNS}
Our evaluation within the unified framework reveals that current learning-based methods do not exhibit a clear advantage over rule-based strategies in terms of time efficiency or improvement capacity. This is primarily due to high time overheads, inaccurate predictions, or the use of inappropriate algorithms in these methods. Nevertheless, our comprehensive analysis indicates several promising future directions for improving MAPF-LNS.

\noindent\textbf{1) Properly targeting high-delayed agents.} The core idea of RandomWalk is to optimize high-delayed agents in each iteration. The superior performance of RandomWalk and RandomWalkProb over others suggests that improving high-delayed agents is an efficient empirical heuristic. Intuitively, focusing on high-delayed agents aligns with the principles of greedy algorithms, which are widely recognized as powerful tools in combinatorial optimization~\cite{papadimitriou1998combinatorial}. They provide efficient solutions to complex problems by making locally optimal choices at each step and strike a balance between solution quality and efficiency, especially for NP-hard problems. Different algorithm designs of RandomWalk and RandomWalkProb result in marginally different performance in experiments, other viable approaches for targeting high-delayed agents can be explored.

\noindent\textbf{2) Contextual bandit for sequential decision-making.} We observe that the supervised learning methods, e.g., SVM and NNS, incur high time overheads. In contrast, Bandit alternates among rule-based strategies in each iteration as sequential decision-making with minimal computational overhead. This makes Bandit particularly suitable for integration into the MAPF-LNS framework. Although Bandit employs an inappropriate non-contextual bandit algorithm, it performs well in two domains under a $60$s time limit. This suggests that a well learned policy for selecting the best rule-based strategies at each time step can significantly enhance MAPF-LNS. Therefore, exploring contextual bandit algorithms is a promising direction. Contextual bandits can address the theoretical limitations of non-contextual approaches by incorporating contextual information into decision-making, potentially leading to better and robust empirical results.

\begin{figure}[t!]
\begin{center}
\includegraphics[width=0.49\textwidth]{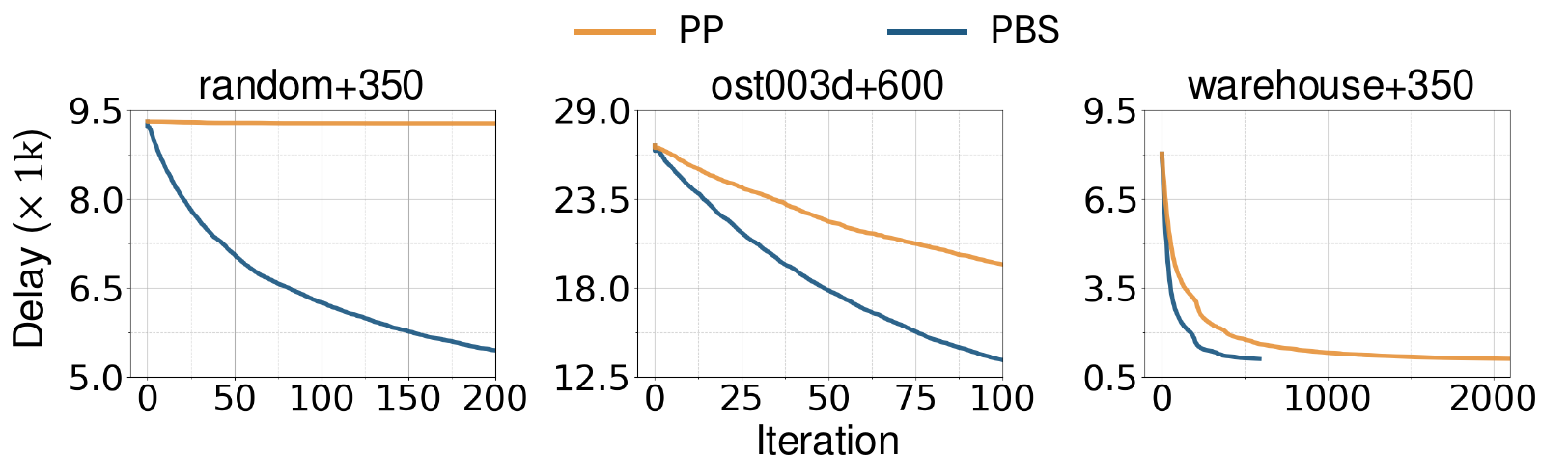}
\end{center}
\vspace{-13pt}
\caption{\textbf{Delay vs. Iteration when the replan solver is PBS or PP}. The neighborhood selection strategy is RandomWalk with a neighborhood size of 25 in all cases.}
\label{fig:delay-iter-pp-pbs}
\vspace{-13pt}
\end{figure}

\noindent\textbf{3) Learning the priority order of replan agents.} We observe that PBS performs better than PP on a per-iteration basis. This is illustrated by the delay-versus-iteration curves shown in Figure~\ref{fig:delay-iter-pp-pbs} for various scenarios. In \texttt{random} and \texttt{ostd003d}, the disparity in delay elimination between PBS and PP is huge. In \texttt{warehouse}, although PP initially reduces delays quickly, it still requires significantly more iterations to achieve the same final delay as PBS. The better per-iteration performance of PBS is due to its strategy of searching for partial priorities among replan agents. In contrast, PP randomly assigns full priority to agents. However, as discussed in Sec.~\ref{sec:pp-vs-pbs}, PBS struggles with time efficiency when evaluated on a runtime basis, as searching for partial priorities is computationally expensive. This highlights an opportunity for improvement: if a fast learning model can predict a reasonable priority order for replan agents, it can enhance time efficiency while improving solution quality.

\begin{table}[b!]

\vspace{-10pt}
\centering
\caption{Final delay of Bandit and Uni-Bandit. Time limit is $300$s.}
\vspace{-10pt}
\label{table:bandit-abliation}
\setlength{\tabcolsep}{2pt}
\resizebox{0.45\textwidth}{!}{
\begin{tabular}{@{}c|c|rr|rr|c|c|rr|rr@{}}
\toprule 
\multirow{2}{*}{} & \multirow{2}{*}{n}  & \multicolumn{2}{c|}{Delay} & \multicolumn{2}{c|}{AUC} &  \multirow{2}{*}{} & \multirow{2}{*}{n} & \multicolumn{2}{c|}{Delay} & \multicolumn{2}{c}{AUC} \\ 
\cmidrule{3-6} \cmidrule{9-12}
 &    & Bandit     & Uni-B    & Bandit    & Uni-B   &   &    & Bandit     & Uni-B    & Bandit    & Uni-B   \\ \midrule
\multicolumn{1}{c|}{\multirow{5}{*}{\rotatebox{90}{empty}}}     & 300 & \text{386.3}      & \textcolor{black}{391.7}         & \text{13.3}      & \textcolor{black}{13.6}         & \multicolumn{1}{c|}{\multirow{5}{*}{\rotatebox{90}{random}}}  & 150 & \text{330.1}      & \textcolor{black}{330.5}         & \textcolor{black}{10.3}      & \text{10.3}         \\
\multicolumn{1}{c|}{}                           & 350 & \text{811.5}      & \textcolor{black}{812.4}         & \text{28.5}      & \textcolor{black}{28.5}         & \multicolumn{1}{c|}{}                         & 200 & \textcolor{black}{779.1}      & \text{778.7}         & \textcolor{black}{24.9}      & \text{24.8}         \\
\multicolumn{1}{c|}{}                           & 400 & \text{1537.2}     & \textcolor{black}{1547.1}        & \text{54.2}      & \textcolor{black}{54.7}         & \multicolumn{1}{c|}{}                         & 250 & \text{1507.3}     & \textcolor{black}{1525.7}        & \text{49.3}      & \textcolor{black}{49.7}         \\
\multicolumn{1}{c|}{}                           & 450 & \text{2753.7}     & \textcolor{black}{2761.2}        & \textcolor{black}{96.5}      & \text{96.0}         & \multicolumn{1}{c|}{}                         & 300 & \text{2746.0}     & \textcolor{black}{2760.5}        & \text{92.0}      & \textcolor{black}{92.0}         \\
\multicolumn{1}{c|}{}                           & 500 & \textcolor{black}{4318.5}     & \text{4302.1}        & \textcolor{black}{150.8}     & \text{149.5}        & \multicolumn{1}{c|}{}                         & 350 & \text{4564.1}     & \textcolor{black}{4565.1}        & \text{155.9}     & \textcolor{black}{155.1}        \\ \midrule
\multicolumn{1}{c|}{\multirow{5}{*}{\rotatebox{90}{warehouse}}} & 150 & \text{107.9}      & \textcolor{black}{111.8}         & \text{3.6}       & \textcolor{black}{3.7}          & \multicolumn{1}{c|}{\multirow{5}{*}{\rotatebox{90}{ost003d}}} & 200 & \text{158.2}      & \textcolor{black}{163.8}         & \textcolor{black}{8.3}       & \text{8.3}          \\
\multicolumn{1}{c|}{}                           & 200 & \textcolor{black}{239.4}      & \text{234.3}         & \textcolor{black}{8.0}       & \text{7.9}          & \multicolumn{1}{c|}{}                         & 300 & \textcolor{black}{532.8}      & \text{484.3}         & \textcolor{black}{33.4}      & \text{31.2}         \\
\multicolumn{1}{c|}{}                           & 250 & \textcolor{black}{414.2}      & \text{414.0}         & \text{14.2}      & \textcolor{black}{14.5}         & \multicolumn{1}{c|}{}                         & 400 & \text{1276.3}     & \textcolor{black}{1327.3}        & \textcolor{black}{75.2}      & \text{75.0}         \\
\multicolumn{1}{c|}{}                           & 300 & \text{669.5}      & \textcolor{black}{677.2}         & \text{23.8}      & \textcolor{black}{24.1}         & \multicolumn{1}{c|}{}                         & 500 & \textcolor{black}{3059.8}     & \text{2873.5}        & \textcolor{black}{163.6}     & \text{155.8}        \\
\multicolumn{1}{c|}{}                           & 350 & \textcolor{black}{1047.7}     & \text{1042.6}        & \textcolor{black}{38.2}      & \text{39.1}         & \multicolumn{1}{c|}{}                         & 600 & \text{6093.7}     & \textcolor{black}{6430.8}        & \text{311.4}     & \textcolor{black}{321.9}        \\ \midrule
\multicolumn{1}{c|}{\multirow{5}{*}{\rotatebox{90}{den520d}}}   & 500 & \textcolor{black}{607.8}      & \text{593.6}         & \textcolor{black}{47.8}      & \text{47.3}         & \multicolumn{1}{c|}{\multirow{5}{*}{\rotatebox{90}{Paris}}}   & 350 & \text{71.9}       & \textcolor{black}{74.0}          & \text{4.6}       & \textcolor{black}{5.0}          \\
\multicolumn{1}{c|}{}                           & 600 & \textcolor{black}{1247.0}     & \text{1234.2}        & \textcolor{black}{89.2}      & \text{87.9}         & \multicolumn{1}{c|}{}                         & 450 & \textcolor{black}{130.8}      & \text{120.8}         & \textcolor{black}{11.0}      & \text{10.1}         \\
\multicolumn{1}{c|}{}                           & 700 & \textcolor{black}{2297.4}     & \text{2195.9}        & \textcolor{black}{150.2}     & \text{144.2}        & \multicolumn{1}{c|}{}                         & 550 & \text{205.3}      & \textcolor{black}{212.5}         & \text{20.7}      & \textcolor{black}{21.4}         \\
\multicolumn{1}{c|}{}                           & 800 & \text{3330.3}     & \textcolor{black}{3607.8}        & \text{209.4}     & \textcolor{black}{221.5}        & \multicolumn{1}{c|}{}                         & 650 & \textcolor{black}{307.5}      & \text{303.0}         & \textcolor{black}{37.0}      & \text{36.9}         \\
\multicolumn{1}{c|}{}                           & 900 & \text{5343.5}     & \textcolor{black}{5421.0}        & \text{308.9}     & \textcolor{black}{310.6}        & \multicolumn{1}{c|}{}                         & 750 & \textcolor{black}{577.0}      & \text{523.6}         & \textcolor{black}{70.6}      & \text{66.2}         \\ \bottomrule
\end{tabular}
}

\vspace{1pt}

\parbox{0.45\textwidth}{\scriptsize Note: `Uni-B' represents the baseline Uni-Bandit, which uniformly selects a neighborhood size from ${2, 4, 8, 16, 32}$ at random.} 
\vspace{-10pt}
\end{table}

\begin{figure}[t!]
\vspace{-2pt}
\centering
\includegraphics[width=0.48\textwidth]{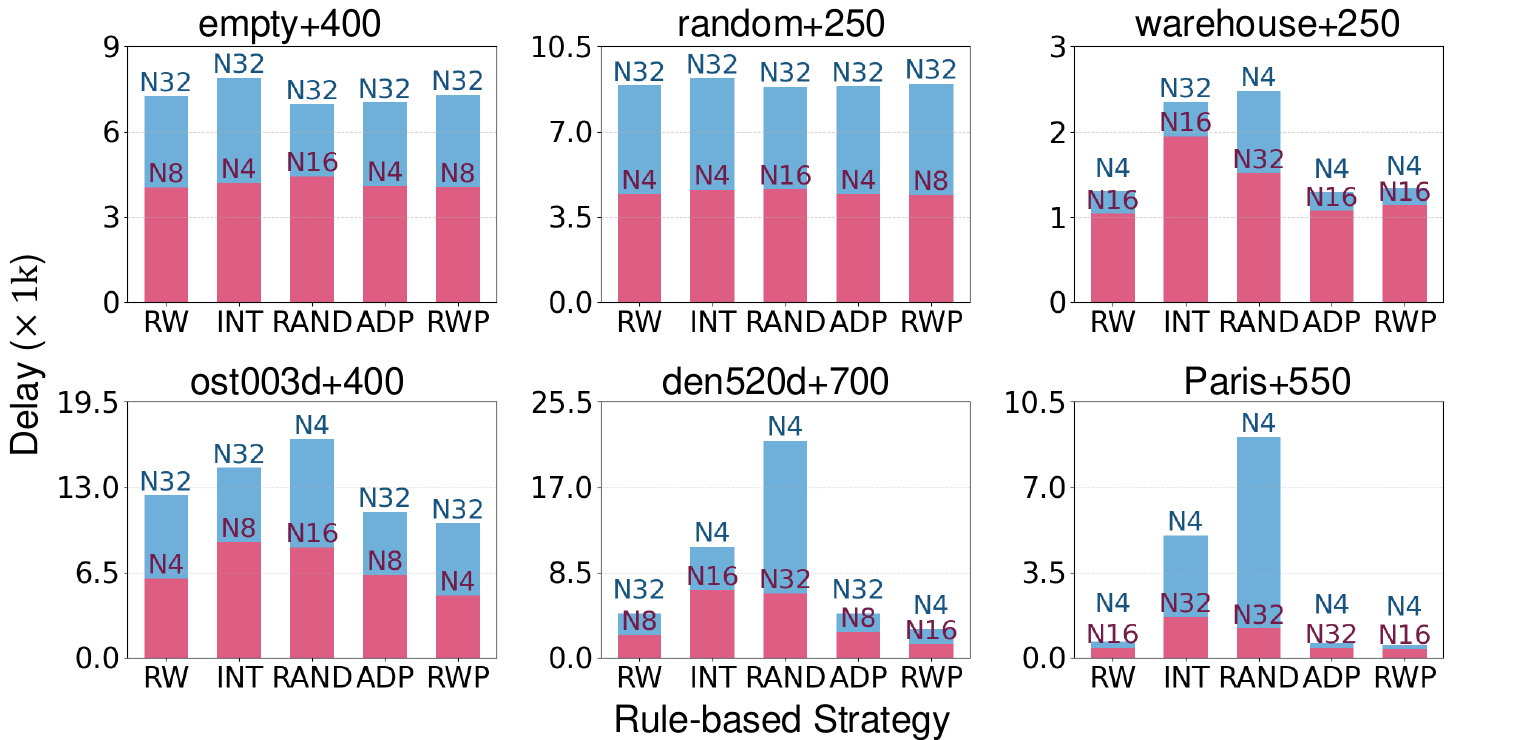}
\caption{\textbf{Final Delays using the Best and Worst neighborhood size within $300$s}. RW, INT, RAND, ADP, and RWP stand for RandomWalk, Intersection, Random, Adaptive, and RandomWalkProb. The blue and pink columns indicate the highest and lowest delays. The neighborhood sizes are labeled at the top of each subfigure.}
\label{fig:nbsize-best-worst}

\vspace{-10pt}

\end{figure}

\noindent\textbf{4) Identifying the suitable neighborhood size.}  
Neighborhood size is a critical factor for MAPF-LNS, but is underexplored in previous papers. Intuitively, a smaller neighborhood size allows for faster iterations but may limit the improvement in solution quality. Conversely, a larger neighborhood size can lead to more improvement per iteration, but at the cost of increased computational time. Thus, there is a trade-off between runtime efficiency and the improvement quality. To highlight the importance of neighborhood size, we compare the final delays achieved by rule-based strategies using the best and worst neighborhood sizes within a $300$s time limit, as shown in Fig.~\ref{fig:nbsize-best-worst}. Since a neighborhood size of $2$ is generally ineffective, we consider sizes from $\{4,8,16,32\}$. In most cases, using the best size reduces final delays by approximately $50\%$ compared to the least favorable size. This difference suggests the potential for performance gains by learning to identify a suitable neighborhood size dynamically.

The only work that attempts to dynamically determine neighborhood size is Bandit. However, as previously discussed, it employs a non-contextual algorithm, which is unsuitable for this purpose. We observe that its performance is similar to selecting neighborhood sizes uniformly at random (we build a baseline, named Uni-Bandit, by modifying the second arm in Bandit to randomly choose neighborhood sizes). The comparison in Table~\ref{table:bandit-abliation} validates this similarity. Therefore, a more sophisticated approach, such as a contextual bandit or other advanced methods, is necessary for effective neighborhood size selection.

\vspace{-0.03in}
\section{Conclusion}
\vspace{-0.03in}
In this work, we conducted a comprehensive reevaluation of prominent MAPF-LNS methods, including recent advances leveraging machine learning. We identified several pitfalls in their evaluations and proposed a unified framework to address these challenges. Our results demonstrate that current learning-based methods fail to exhibit a clear advantage over simple rule-based heuristics, while RandomWalk and its variant RandomWalkProb, consistently deliver robust performance across diverse scenarios. Furthermore, our evaluation and extensive experiments highlight promising directions for advancing MAPF-LNS, such as targeting high-delayed agents, employing contextual algorithms for strategy selection, learning replanning agent orders, and dynamically identifying suitable neighborhood sizes. We believe this work will encourage future research to adopt more rigorous experimental designs and inspire innovative approaches to enhancing MAPF-LNS through machine learning.

\bibliography{aaai25}

\clearpage

\section*{APPENDIX}

\section{Summary of Unified Evaluation and RandomWalkProb}
We provide a summary of the unified evaluation framework, including details on initial solutions, replan solvers, and the number of agents in each map. Additionally, we clarify the missing implementation details of RandomWalkProb, a modified variant of RandomWalk.

\subsection{Unified Evaluation}

The initial solutions, replan solvers, and neighborhood sizes used for different methods, and the number of agents evaluated in different maps are summarized in Table~\ref{table:diff-eval-setting}. All methods are evaluated on a machine equipped with an Intel E5-2683 CPU with a memory limit of 2G. Specifically, the execution of the neural network in Neural-LNS is performed on an NVIDIA P100 GPU.


\begin{table}[h]
\caption{\textbf{Top}: Initial solutions, replan solvers, neighborhood sizes used for evaluating different methods. 'Ori-' refers to methods evaluated according to its original papers. \textbf{Bottom}: number of agents evaluated in different maps.}
\label{table:diff-eval-setting}
\centering
\resizebox{0.465\textwidth}{!}{
{\small
\begin{tabular}{c|c|c|c}
\toprule
Method &  Initial Solution  &  Replan Solver & Neighborhood size\\
\midrule
RandomWalk   &  LaCAM2, LNS2  & PP  & $\{4,8,16,32\}$ \\
Intersection &  LaCAM2, LNS2  & PP  & $\{4,8,16,32\}$ \\
Random       &  LaCAM2, LNS2  & PP  & $\{4,8,16,32\}$ \\
Adaptive     &  LaCAM2, LNS2  & PP  & $\{4,8,16,32\}$ \\
name         &  LaCAM2, LNS2  & PP  & $\{4,8,16,32\}$ \\
SVM  &  LaCAM2, LNS2  & PP  & $\{4,8,16,32\}$ \\
NNS& LaCAM2, LNS2 & PP  & $\{4,8,16,32\}$ \\
Bandit& LaCAM2, LNS2  & PP  &  $\{2,4,8,16,32\}$ by second arm\\
\midrule
Ori-SVM &PP, PPS, EECBS&PP & Uniformly select from 5 to 16 \label{sup:Ori-SVM-setting-row} \\
Ori-NNS& PP, PPS  & PBS & $\{10,25,50\}$ for different maps\\
Ori-Bandit& PP, PPS, EECBS & PP  &  $\{2,4,8,16,32\}$ by second arm\\

\bottomrule

\end{tabular}
}
}

\vspace{10pt}
\resizebox{0.465\textwidth}{!}{
{\small
\begin{tabular}{c|c}

\toprule
Map &  Number of agents                              \\
\midrule
empty-32-32 (empty) & 300, 350, 400, 450, 500 \\
random-32-32-20 (random) & 150, 200, 250, 300, 350 \\
warehouse-10-20-10-2-1 (warehouse) & 150, 200, 250, 300, 350 \\
ost003d & 200, 300, 400, 500, 600 \\
den520d  & 500, 600, 700, 800, 900 \\
Paris\_1\_256 (Paris) & 350, 450, 550, 650, 750 \\
\bottomrule
\end{tabular}
}
}
\vspace{-10pt}
\end{table}



\subsection{Algorithm Design of RandomWalkProb}
\label{sec:randomwalkprob}
RandomWalkProb shares the same motivation as RandomWalk, which focuses on improving the high-delayed agents. 
However, the original algorithm of RandomWalk has two potential limitations: 1) it uses a tracking set to record previously selected high-delayed agents, preventing them from being selected again for subsequent \texttt{random\_walk} searches even if they still have significant delays; 2) when the neighborhood size is not reached after a \texttt{random\_walk} search, a random agent is chosen as the next starting agent, which is less informative regarding the delays.
Therefore, we conduct two modifications on RandomWalk. First, we remove the tracking set. Second, whenever to start a \texttt{random\_walk} search, the starting agent is sampled according to a probability proportional to its delay, i.e.  $\mathrm{Pr}(a_i)=\frac{delay(p_i)}{\sum_j delay(p_j)}$.

Algo.~\ref{algo:heuristic} is the pseudo code comparing RandomWalk and RandomWalkProb (\rw{gray} lines only exist in \rw{RandomWalk}, and \rwp{blue} lines only exist in \rwp{RandomWalkProb}). RandomWalk tracks previously selected starting agents into a \textit{tabuList} [Line~\ref{rw:1} to Line~\ref{rw:2}], which is reset after all delayed agents have been chosen as the starting agents [Line~\ref{rw:3}]. When the neighborhood size is not reached after one search, a random agent is chosen from $\tilde{A}$ as the start agent [Line~\ref{rw:4}] to perform another \texttt{random\_walk} search. RandomWalkProb selects the starting agent \(a_k\) by sampling according to a probability proportional to delays [Line~\ref{rwp:1} and Line~\ref{rwp:2}].

After selecting the starting agent $a_k$, both RandomWalk and  RandomWalkProb use the same \texttt{random\_walk()} search function [Line 7; Line 10 to Line~\ref{algo:rw-end}] to prioritize moving the starting agent $a_k$ towards a shorter path. The \texttt{random\_walk()} function first randomly selects a start state $x$ along agent $a_k$'s path $p_k$ [Line 11]. It then collects possible vertices $v \in N_x$ where agent $a_k$ at time step $t + 1$ can reach and move towards a shorter path to $g_k$ while ignoring other agents [Line 12]. Since any path passing through vertex $v$ at time step $t+1$ must be at least $t+1 + d(v, g_k)$ in length, the condition for selecting $v$ is $t+1 + d(v, g_k) < l(p_k)$. As long as $N_x$ is not empty and collected agents are not enough [Line 13], the agent moves to a random vertex $y \in N_x$ [Line 14] and adds any agents who collide with action to $\tilde{A}$ [Line 15] and updates the set $N_x$ [Line 17].

\begin{algorithm}[t!]
\caption{\rw{RandomWalk} / \rwp{RandomWalkProb}}\label{algo:heuristic}
\DontPrintSemicolon
\KwIn{Graph $G=(V,E)$, agents $A=\{a_1,...,a_n\}$, neighborhood size $M$, paths $P=\{p_1,...,p_n\}$, \rw{ $tabuList$ from previous LNS iteration, number of delayed agents $N_{delay}$}} 

\rwp{$a_k\sim \mathrm{Pr}(a_k)=\frac{delay(p_k)}{\sum_i delay(p_i)}$\;} \label{rwp:1}

\rw{$a_k \leftarrow \arg \max _{a_i \in A \backslash \text { tabuList }}\left\{\operatorname{delay}\left(p_i\right)\right\}$ \;} \label{rw:1}

\rw{tabuList $\leftarrow$ tabuList $\cup\left\{a_k\right\}$ \;} \label{rw:2}

\rw{\lIf{$|tabuList| = N_{delay}$}{$tabuList \leftarrow \emptyset$}} \label{rw:3}

$~\tilde{A}\leftarrow \{a_k\}$\;

\While{$|\tilde{A}|<M$}{
    $\tilde{A}\leftarrow$ random\_walk($G,a_k,P,\tilde{A},M$)\; 
    \rwp{$a_k\sim \mathrm{Pr}(a_k)=\frac{delay(p_k)}{\sum_i delay(p_i)}$ \;} \label{rwp:2}
    \rw{$a_k \leftarrow$ random agent in $\tilde{A}$ \;} \label{rw:4}
    
}

\SetKwFunction{Rw}{random\_walk} \label{algo:rw-start}
\SetKwProg{Fn}{Function}{:}{}
\Fn{\Rw{$G,a_k,P,\tilde{A},M$}}{
    $(x, t) \leftarrow\left(p_k[t], t\right)$, where $t$ is a random timestep of $p_k$ \;
    $N_x\leftarrow\{v\in V|(x,v)\in E\cup \{(x,x)\}\wedge t+1+d(v,g_k)<l(p_k)\}$\;
    \While{$|N_x|>0\wedge|\tilde{A}|<M$}{
        $y\leftarrow$ a random vertex in $N_x$\;
        $\tilde{A}\leftarrow\tilde{A}\cup\{$ agents collide with action 'moving to $y$' $\}$\;
        $(x,t)\leftarrow (y,t+1)$\;
        $N_x\leftarrow\{v\in V|(x,v)\in E\cup \{(x,x)\}\wedge t+1+d(v,g_k)<l(p_k)\}$\;
    }
    \Return $\tilde{A}$\; \label{algo:rw-end}
}
\vspace{-5pt}
\end{algorithm}


\section{SVM/Neural-LNS Training Details}
\label{sec:svm-nns-train}
We outline the training details for SVM-LNS and Neural-LNS, including replicating their reported performance by adhering to the settings specified in their original papers, and the reevaluation under our unified settings. 

\subsection{SVM-LNS}


\paragraph{Training Data for original SVM-LNS according to \citet{huang2022anytime}.} We use the suggested number of agents in~\citet{huang2022anytime} to train SVM if the map exists in the original paper (i.e. 100 for \texttt{warehouse}, 100 for \texttt{ostd003d}, 200 for \texttt{den520d} and 250 for \texttt{Paris}). For maps not evaluated in \citet{huang2022anytime}, we use 300 agents for \texttt{empty-32-32} and 150 agents for \texttt{random-32-32-20}. The neighborhood size is uniformly selected between 5 and 16 (as denoted in Table~\ref{table:diff-eval-setting}). Following the original paper, we run 16 scenes on each map and switch between RandomWalk and Intersection with equal probability to generated 20 neighborhood candidates. The ground truth ranking information for these 20 candidates is determined by the delay improvement if each neighborhood is removed and replaced.

\paragraph{Training Data for SVM-LNS under our unified settings.} We use the best rule-based strategy with the best neighborhood size for each map to collect training data. The maps contain a medium number of agents. The used strategy with the neighborhood size for collecting training data are the same as training Neural-LNS under our unified setting, e.g., see the first three columns of the bottom part of Table~\ref{table:diff-nns-setting}.

\paragraph{Validation Data.} We run SVM-LNS on 4 additional scenes for each map with 100 iterations to collect validation data. In each iteration, the best neighborhood is selected as the ground truth.

\paragraph{Training.} SVM-LNS trains its SVM model dynamically during execution. The implementation of SVM is using $\mathrm{SVM}^{rank}$\footnote{https://www.cs.cornell.edu/people/tj/svm\_light/svm\_rank.html}, which is suggested by original authors. The model is updated immediately after collecting new data and is then used to gather additional data in the next iteration. Training is conducted on 16 scenes per map, resulting in 16 new data points collected per LNS iteration. The model undergoes training for 100 iterations, with the best model selected based on its average rank on the validation set.

\subsection{Neural-LNS}



\paragraph{Training Data for original Neural-LNS according to \citet{yan2024neural}.} We used the number of agents, rule-based strategies and neighborhood size suggested by the authors to collect data for the training set, which are summarized in the top of Table~\ref{table:diff-nns-setting}.
We run 25 to 50 iterations to collect data for each map until there is no further decrease in delays. 

In each iteration, 100 neighborhood candidates are proposed using the suggested rule-based strategy and neighborhood size. The ground truth ranking of these 100 neighborhood candidates is determined by the delay improvement. 


\begin{table}[t]
\caption{\textbf{Top}: Training Data Collection Strategy of Ori-NNS. \textbf{Bottom}: Training Data Collection Strategy of NNS trained under our unified setting. `NB' is the neighborhood size.}
\label{table:diff-nns-setting}
\centering
\resizebox{0.465\textwidth}{!}{
\begin{tabular}{c|c|c|c|c|c}
\toprule
\multicolumn{6}{c}{Ori-NNS Training Data Collection} \\ 
\midrule
Map       & Strategy   & NB & Iteration & Scene & Data Amount \\ \midrule
empty     & Random     & 50 & 50        & 5000   & 250000 \\ 
random    & RandomWalk & 25 & 50        & 5000   & 250000 \\ 
warehouse & RandomWalk & 25 & 25        & 5000   & 125000 \\ 
ost003d   & RandomWalk & 10 & 25        & 1000   & 25000  \\ 
den520d   & RandomWalk & 25 & 50        & 5000   & 250000 \\ 
Paris     & RandomWalk & 25 & 50        & 4000   & 200000 \\ 
\bottomrule

\end{tabular}

}

\vspace{10pt}
\resizebox{0.465\textwidth}{!}{

\begin{tabular}{c|c|c|c|c|c}

\toprule
\multicolumn{6}{c}{NNS Training Data Collection} \\
\midrule
Map       & Strategy         & NB & Iteration & Scenes & Data Amount \\ \midrule
empty     & Adaptive         & 8  & 1400      & 300    & 420000  \\ 
random    & RandomWalk       & 8  & 1000      & 100    & 100000  \\ 
warehouse & Adaptive         & 32 & 200       & 200    & 40000   \\ 
ost003d   & RandomWalkProb   & 16 & 400       & 250    & 100000  \\ 
den520d   & RandomWalkProb   & 16 & 500       & 200    & 100000  \\ 
Paris     & RandomWalkProb   & 32 & 200       & 350    & 70000   \\ 
\bottomrule
\end{tabular}
}
\end{table}

\paragraph{Training Data for Neural-LNS under our unified settings.} We use the best rule-based strategy with the best neighborhood size for each map to collect training data. The maps contain a medium number of agents.  The replan solver is PP, which requires more iterations than PBS to converge. As a result, we use fewer scenes per map to gather a comparable amount of data. The exact number of iterations, scenes, neighborhood sizes, and removal strategies for each map are detailed in the bottom part of Table~\ref{table:diff-nns-setting}.

\paragraph{Validation Data.} We run additional 25 scenes to gather validation data for original NNS. We run additional 4 scenes to collect validation data for NNS under our unified setting. Similar to collecting training data,  we use fewer scenes than original NNS because we use PP as the replan solver, allowing more iterations to generate more data from a single scene.



\paragraph{Training.} For each map, the model is trained on the corresponding training set. We stop the training when the loss converges and the average ranking on the validation set no longer improves for another $1,000$ steps. We calculate the average rank on the validation set to select the best model checkpoint for inference. Here, 'average ranking' means the mean ranking of the best neighborhood predicted by the model appearing in the ground truth ranking over the validation dataset.

\paragraph{Hyperparameters.} We search for the optimal learning rate within $\left\{0.1, 0.01, 0.001, 0.0001, 0.00001\right\}$ and choose $0.00001$. This is smaller than the $0.0001$ learning rate used in the original paper. We find that a smaller learning rate results in a more stable reduction in loss on our training data. We use a batch size of $16$ and train the model for $10,000$ to $100,000$ steps until the loss and validation score no longer improve for an additional $1,000$ steps. The entire training process is completed in less than $24$ hours. 

\paragraph{Testing.} We use an NVIDIA P100 GPU for neural network inference. The average GPU inference overhead of is summarized in Table~\ref{table:Orig-Neural-Overhead}.

\begin{table}[h!]
\centering
\caption{Average Overhead of NNS inference on GPU.}
\label{table:Orig-Neural-Overhead}
\resizebox{0.465\textwidth}{!}{
\begin{tabular}{c|c|c|c|c|c|c|c|c}
\toprule
\multicolumn{1}{l|}{Map}         & N & Overhead (s) & \multicolumn{1}{l|}{Map}             & N & Overhead (s) & \multicolumn{1}{l|}{Map}                    & N & Overhead (s)  \\ \midrule
\multirow{5}{*}{\rotatebox{90}{empty}} & 300      & 0.016   & \multirow{5}{*}{\rotatebox{90}{random}} & 150      & 0.014   & \multirow{5}{*}{\rotatebox{90}{warehouse}} & 150      & 0.042    \\
                                 & 350      & 0.017    &                                      & 200      & 0.015   &                                             & 200      & 0.043   \\
                                 & 400      & 0.019   &                                      & 250      & 0.021   &                                             & 250      & 0.043   \\
                                 & 450      & 0.022   &                                      & 300      & 0.024   &                                             & 300      & 0.044   \\
                                 & 500      & 0.028   &                                      & 350      & 0.026   &                                             & 350      & 0.044   \\ \toprule
\multirow{5}{*}{\rotatebox{90}{ost003d}}     & 200      & 0.020   & \multirow{5}{*}{\rotatebox{90}{den520d}}         & 500      & 0.033   & \multirow{5}{*}{\rotatebox{90}{Paris}}            & 350      & 0.038   \\
                                 & 300      & 0.020   &                                      & 600      & 0.041   &                                             & 450      & 0.038   \\
                                 & 400      & 0.024   &                                      & 700      & 0.041   &                                             & 550      & 0.038   \\
                                 & 500      & 0.024   &                                      & 800      & 0.055   &                                             & 650      & 0.040   \\
                                 & 600      & 0.027   &                                      & 900      & 0.045   &                                             & 750      & 0.042   \\ \toprule
\end{tabular}
}
\end{table}

\section{Additional Results}
\subsection{PP vs. PBS}

PBS is claimed to be superior over PP in some cases by \citet{yan2024neural} (e.g., see Table 1 of~\citep{yan2024neural}). We investigate the efficiency of these two replan solvers in all 30 evaluation cases, i.e., 6 maps with 5 different agent amounts, along with two initial solvers. We fix the neighborhood selection heuristic as RandomWalk and the neighborhood size as 25, which are suggested by \citet{yan2024neural} in most cases. We report the total iterations, final delays, and  AUC of the delay-versus-time curves with time limits $60$s and $300$s in Table~\ref{table:full-pp-pbs-300s-60s}. The results where PBS is better than PP is highlighted in red. For final delays, PP is better than PBS in $72.5\%$ (87/120) cases. For AUC, PP is better than PBS in $81.7\%$ (98 / 120) cases. Even though PBS is better than PP in \texttt{random} map, the final delays and AUC are relatively close. In general, PP runs significantly faster than PBS  and thus can explore a substantially larger number of neighborhoods within the time limit.

\subsection{Full Results}
\label{sec:full-results}
The complete results are shown in 
Table~\ref{table:300s-lns2} (time limit: $300$s, initial solver: LNS2), Table~\ref{table:60s-lns2} (time limit: $60$s, initial solver: LNS2), Table~\ref{table:300s-lacam2} (time limit: $300$s, initial solver: LaCAM2), and Table~\ref{table:60s-lacam2} (time limit: $60$s, initial solver: LaCAM2).

\begin{table}[ht]
    \caption{Total iterations, final delays, and AUC in different evaluation cases within a time limit of \textbf{60s} and \textbf{300s}, using PP and PBS as replan solvers. The neighborhood selection strategy is RandomWalk with a neighborhood size of 25. 'In' refers to the algorithm used for finding initial solutions. For both final delays and AUC, lower values are better. The settings where PBS performs better are highlighted in red, for all other settings, PP is superior. } 
    \label{table:full-pp-pbs-300s-60s}
    \centering
    \setlength{\tabcolsep}{2pt}
    \resizebox{0.485\textwidth}{!}{
    \begin{tabular}{@{}|c|c|c|cc|rr|rr|c|c|c|cc|rr|rr|@{}}
    \toprule
    \multicolumn{18}{|c|}{\textbf{Run Time Limit: 60s}} \\ 
    \midrule
    \multirow{3}{*}{} & \multirow{3}{*}{In} & \multirow{3}{*}{n} & \multicolumn{2}{c|}{Iter (x1k)} & \multicolumn{2}{c|}{Final delays} & \multicolumn{2}{c|}{AUC (x10k)} & \multirow{3}{*}{} & \multirow{3}{*}{In} & \multirow{3}{*}{n} & \multicolumn{2}{c|}{Iter (x1k)} & \multicolumn{2}{c|}{Final delays} & \multicolumn{2}{c|}{AUC (x10k)} \\
    \cmidrule(lr){4-9} \cmidrule(lr){13-18}
    & & & PP & PBS & \multicolumn{1}{c}{PP} & \multicolumn{1}{c|}{PBS} & \multicolumn{1}{c}{PP} & \multicolumn{1}{c|}{PBS} & & & & PP & PBS & \multicolumn{1}{c}{PP} & \multicolumn{1}{c|}{PBS} & \multicolumn{1}{c}{PP} & \multicolumn{1}{c|}{PBS} \\
    \midrule
    \multirow{10}{*}{\rotatebox{90}{empty}} & \multirow{5}{*}{\rotatebox{90}{LaCAM2}} & 300 & 8.21 & 0.48 & 439.6 & \textcolor{red}{424.5} & 3.9 & 6.07 & \multirow{10}{*}{\rotatebox{90}{random}} & \multirow{5}{*}{\rotatebox{90}{LaCAM2}} & 150 & 6.21 & 0.58 & 352.1 & \textcolor{red}{343.1} & 2.5 & 3.6 \\
     &  & 350 & 4.13 & 0.22 & 1,127.6 & 1,286.8 & 9.4 & 14.2 &  &  & 200 & 3.01 & 0.20 & 952.5 & \textcolor{red}{874.9} & 7.3 & 7.6 \\
     &  & 400 & 2.24 & 0.12 & 2,663.1 & 2,982.4 & 20.2 & 27.1 &  &  & 250 & 2.22 & 0.07 & 2,449.5 & \textcolor{red}{2,443.6} & 18.6 & 21.7 \\
     &  & 450 & 2.15 & 0.11 & 5,110.7 & 5,211.0 & 36.9 & 41.4 &  &  & 300 & 2.03 & 0.04 & 5,318.2 & 5,693.2 & 39.6 & 44.9 \\
     &  & 500 & 2.40 & 0.08 & 8,400.6 & 8,815.3 & 59.2 & 64.5 &  &  & 350 & 2.61 & 0.00 & 14,729.1 & \textcolor{red}{14,630.2} & 82.4 & 83.7 \\
    \cmidrule(lr){2-9} \cmidrule(lr){11-18}
    & \multirow{5}{*}{\rotatebox{90}{LNS2}} & 300 & 8.98 & 0.44 & 431.7 & 436.9 & 3.5 & 4.8 & & \multirow{5}{*}{\rotatebox{90}{LNS2}} & 150 & 7.39 & 0.63 & 350.1 & \textcolor{red}{346.9} & 2.3 & 2.8 \\
     &  & 350 & 4.22 & 0.25 & 1,109.8 & \textcolor{red}{1,081.8} & 8.7 & 9.8 &  &  & 200 & 2.88 & 0.19 & 959.6 & \textcolor{red}{875.5} & 6.8 & \textcolor{red}{6.5} \\
     &  & 400 & 2.28 & 0.15 & 2,570.1 & \textcolor{red}{2,238.2} & 18.7 & \textcolor{red}{17.5} &  &  & 250 & 1.99 & 0.05 & 2,423.8 & \textcolor{red}{2,301.4} & 16.7 & \textcolor{red}{16.2} \\
     &  & 450 & 1.83 & 0.09 & 4,873.5 & \textcolor{red}{4,293.6} & 32.7 & \textcolor{red}{29.9} &  &  & 300 & 1.47 & 0.03 & 5,309.6 & \textcolor{red}{4,533.1} & 33.7 & \textcolor{red}{30.4} \\
     &  & 500 & 1.51 & 0.05 & 7,817.6 & \textcolor{red}{6,874.2} & 49.3 & \textcolor{red}{45.2} &  &  & 350 & 1.57 & 0.02 & 8,966.9 & \textcolor{red}{8,076.5} & 54.7 & \textcolor{red}{51.2} \\
    \midrule
    \multirow{10}{*}{\rotatebox{90}{warehouse}} & \multirow{5}{*}{\rotatebox{90}{LaCAM2}} & 150 & 6.41 & 0.60 & 116.8 & 133.1 & 1.4 & 6.6 & \multirow{10}{*}{\rotatebox{90}{ost003d}} & \multirow{5}{*}{\rotatebox{90}{LaCAM2}} & 200 & 1.73 & 0.09 & 198.5 & 1,074.5 & 3.7 & 12.4 \\
     &  & 200 & 2.94 & 0.25 & 259.8 & 319.3 & 7.3 & 7.6 &  &  & 300 & 0.93 & 0.07 & 988.8 & 3,117.1 & 13.9 & 35.6 \\
     &  & 250 & 1.74 & 0.13 & 486.7 & 941.4 & 18.6 & 21.7 &  &  & 400 & 0.57 & 0.03 & 3,285.9 & 9,320.1 & 37.3 & 81.6 \\
     &  & 300 & 1.06 & 0.07 & 845.2 & 2,987.7 & 39.6 & 44.9 &  &  & 500 & 0.19 & 0.01 & 12,164.9 & 21,539.9 & 98.6 & 145.5 \\
     &  & 350 & 0.65 & 0.03 & 1,625.8 & 7,963.7 & 82.4 & 83.7 &  &  & 600 & 0.13 & 0.01 & 27,290.3 & 35,498.7 & 188.7 & 221.2 \\
    \cmidrule(lr){2-9} \cmidrule(lr){11-18}
    & \multirow{5}{*}{\rotatebox{90}{LNS2}} & 150 & 6.59 & 0.57 & 122.1 & 128.3 & 2.3 & 2.8 & & \multirow{5}{*}{\rotatebox{90}{LNS2}} & 200 & 1.75 & 0.06 & 183.9 & 897.6 & 2.8 & 9.5 \\
     &  & 200 & 2.65 & 0.27 & 266.8 & 310.2 & 6.8 & \textcolor{red}{6.5} &  &  & 300 & 0.92 & 0.02 & 915.5 & 4,630.9 & 12.3 & 35.2 \\
        &  & 250 & 1.83 & 0.15 & 477.6 & 760.3 & 16.7 & \textcolor{red}{16.2} &  &  & 400 & 0.52 & 0.03 & 3,230.7 & 8,032.7 & 32.4 & 62.3 \\
        &  & 300 & 1.15 & 0.09 & 832.7 & 1,740.3 & 33.7 & \textcolor{red}{30.4} &  &  & 500 & 0.21 & 0.01 & 9,335.3 & 16,709.3 & 72.3 & 107.8 \\
        &  & 350 & 0.73 & 0.06 & 1,495.0 & 3,237.5 & 54.7 & \textcolor{red}{51.2} &  &  & 600 & 0.15 & 0.01 & 17,998.3 & 24,525.7 & 125.2 & 152.2 \\
    \midrule
    \multirow{10}{*}{\rotatebox{90}{den520d}} & \multirow{5}{*}{\rotatebox{90}{LaCAM2}} & 500 & 1.44 & 0.04 & 871.6 & 8,082.4 & 18.5 & 77.1 & \multirow{10}{*}{\rotatebox{90}{Paris}} & \multirow{5}{*}{\rotatebox{90}{LaCAM2}} & 350 & 7.48 & 0.16 & 99.8 & 817.8 & 1.8 & 18.5 \\
     &  & 600 & 1.07 & 0.02 & 2,266.5 & 17,753.3 & 35.3 & 129.3 &  &  & 450 & 6.69 & 0.10 & 134.3 & 3,032.1 & 6.2 & 79.9 \\
     &  & 700 & 0.86 & 0.02 & 4,396.1 & 24,979.7 & 57.9 & 175.6 &  &  & 550 & 5.37 & 0.05 & 213.7 & 8,664.9 & 10.6 & 119.2 \\
     &  & 800 & 0.53 & 0.01 & 9,205.8 & 35,921.6 & 96.9 & 234.4 &  &  & 650 & 4.58 & 0.03 & 298.0 & 15,771.8 & 18.5 & 165.3 \\
     &  & 900 & 0.49 & 0.01 & 12,900.4 & 45,686.7 & 124.4 & 291.1 &  &  & 750 & 3.57 & 0.02 & 483.6 & 24,171.6 & 18.5 & 165.3 \\
    \cmidrule(lr){2-9} \cmidrule(lr){11-18}
    & \multirow{5}{*}{\rotatebox{90}{LNS2}} & 500 & 1.28 & 0.05 & 899.6 & 6,195.8 & 15.9 & 52.9 & & \multirow{5}{*}{\rotatebox{90}{LNS2}} & 350 & 5.98 & 0.17 & 82.2 & 383.7 & 1.0 & 8.4 \\
     &  & 600 & 1.72 & 0.06 & 1,321.3 & 8,485.5 & 29.8 & 79.2 &  &  & 450 & 6.44 & 0.11 & 138.7 & 2,274.2 & 1.8 & 22.8 \\
     &  & 700 & 0.78 & 0.02 & 4,436.5 & 16,642.9 & 49.1 & 111.9 &  &  & 550 & 4.72 & 0.06 & 219.3 & 4,878.6 & 4.1 & 46.0 \\
     &  & 800 & 0.61 & 0.02 & 7,342.8 & 21,909.0 & 73.2 & 142.2 &  &  & 650 & 4.49 & 0.04 & 317.1 & 9,304.6 & 4.6 & 14.7 \\
     &  & 900 & 0.44 & 0.01 & 13,032.0 & 29,352.2 & 105.6 & 181.4 &  &  & 750 & 3.07 & 0.03 & 614.9 & 14,707.1 & 14.8 & 104.5 \\
    \bottomrule
    \end{tabular}
    
    }
 
    \resizebox{0.485\textwidth}{!}{

\begin{tabular}
{@{}|c|c|c|cc|rr|rr|c|c|c|cc|rr|rr|@{}}
\toprule
\multicolumn{18}{|c|}{\textbf{Run Time Limit: 300s}} \\ 
\midrule
\multirow{3}{*}{} & \multirow{3}{*}{In} & \multirow{3}{*}{n} & \multicolumn{2}{c|}{Iter (x10k)} & \multicolumn{2}{c|}{Final delays} & \multicolumn{2}{c|}{AUC (x10k)} & \multirow{3}{*}{} & \multirow{3}{*}{In} & \multirow{3}{*}{n} & \multicolumn{2}{c|}{Iter (x10k)} & \multicolumn{2}{c|}{Final delays} & \multicolumn{2}{c|}{AUC (x10k)} \\
\cmidrule(lr){4-9} \cmidrule(lr){13-18}
    & & & PP & PBS & \multicolumn{1}{c}{PP} & \multicolumn{1}{c|}{PBS} & \multicolumn{1}{c}{PP} & \multicolumn{1}{c|}{PBS} & & & & PP & PBS & \multicolumn{1}{c}{PP} & \multicolumn{1}{c|}{PBS} & \multicolumn{1}{c}{PP} & \multicolumn{1}{c|}{PBS}  \\
\midrule
\multirow{10}{*}{\rotatebox{90}{empty}} & \multirow{5}{*}{\rotatebox{90}{LaCAM2}} & 300 & 4.29 & 0.21 & 367.9 & \textcolor{red}{332.3} & 13.3 & 14.6 & \multirow{10}{*}{\rotatebox{90}{random}} & \multirow{5}{*}{\rotatebox{90}{LaCAM2}} & 150 & 3.11 & 0.24 & 329.5 & \textcolor{red}{321.4} & 10.6 & 11.4 \\
 &  & 350 & 2.07 & 0.12 & 840.3 & \textcolor{red}{692.6} & 31.7 & 33.9 &  &  & 200 & 1.58 & 0.10 & 789.8 & \textcolor{red}{723.6} & 27.3 & \textcolor{red}{25.8} \\
 &  & 400 & 1.09 & 0.06 & 1971.6 & \textcolor{red}{1544.3} & 73.2 & 75.0 &  &  & 250 & 1.24 & 0.03 & 1857.1 & \textcolor{red}{1618.0} & 67.8 & \textcolor{red}{66.5} \\
 &  & 450 & 1.07 & 0.04 & 3899.8 & \textcolor{red}{3259.0} & 140.4 & 135.0 &  &  & 300 & 1.21 & 0.02 & 4102.7 & \textcolor{red}{3603.0} & 147.9 & \textcolor{red}{145.2} \\
 &  & 500 & 1.28 & 0.03 & 6615.2 & \textcolor{red}{6092.5} & 232.3 & \textcolor{red}{231.8} &  &  & 350 & 1.06 & 0.01 & 8553.1 & \textcolor{red}{7610.3} & 315.3 & \textcolor{red}{310.0} \\
\cmidrule(lr){2-9} \cmidrule(lr){11-18}
& \multirow{5}{*}{\rotatebox{90}{LNS2}} & 300 & 4.64 & 0.19 & 364.0 & \textcolor{red}{339.0} & 12.6 & 13.5 & & \multirow{5}{*}{\rotatebox{90}{LNS2}} & 150 & 3.78 & 0.23 & 333.9 & \textcolor{red}{328.9} & 10.5 & 10.8 \\
 &  & 350 & 2.09 & 0.12 & 853.1 & \textcolor{red}{695.7} & 31.2 & \textcolor{red}{28.8} &  &  & 200 & 1.53 & 0.10 & 812.8 & \textcolor{red}{741.7} & 27.4 & 25.2 \\
 &  & 400 & 1.10 & 0.07 & 1920.6 & \textcolor{red}{1400.2} & 70.0 & \textcolor{red}{57.5} &  &  & 250 & 1.21 & 0.03 & 1841.2 & \textcolor{red}{1651.3} & 65.4 & 61.2 \\
 &  & 450 & 1.00 & 0.04 & 3749.9 & \textcolor{red}{3010.3} & 132.3 & \textcolor{red}{112.7} &  &  & 300 & 0.95 & 0.01 & 4253.8 & \textcolor{red}{3527.6} & 146.1 & 123.1 \\
 &  & 500 & 0.98 & 0.02 & 6447.5 & \textcolor{red}{5352.5} & 216.7 & \textcolor{red}{186.7} &  &  & 350 & 0.87 & 0.01 & 8225.2 & \textcolor{red}{6477.2} & 260.5 & \textcolor{red}{221.7} \\
\midrule
\multirow{10}{*}{\rotatebox{90}{warehouse}} & \multirow{5}{*}{\rotatebox{90}{LaCAM2}} & 150 & 3.01 & 0.19 & 113.4 & 127.0 & 4.1 & 9.7 & \multirow{10}{*}{\rotatebox{90}{ost003d}} & \multirow{5}{*}{\rotatebox{90}{LaCAM2}} & 200 & 0.87 & 0.04 & 154.0 & 278.8 & 7.6 & 24.6 \\
 &  & 200 & 1.48 & 0.12 & 247.9 & 262.5 & 9.4 & 16.1 &  &  & 300 & 0.50 & 0.03 & 382.0 & 1160.6 & 26.8 & 82.8 \\
 &  & 250 & 0.93 & 0.08 & 435.5 & 445.9 & 18.2 & 38.0 &  &  & 400 & 0.37 & 0.02 & 1009.4 & 4071.9 & 76.5 & 207.7 \\
 &  & 300 & 0.55 & 0.05 & 678.7 & 741.9 & 31.9 & 74.8 &  &  & 500 & 0.13 & 0.01 & 4618.9 & 9758.7 & 275.9 & 483.5 \\
 &  & 350 & 0.35 & 0.03 & 1082.9 & 1471.1 & 61.1 & 154.8 &  &  & 600 & 0.07 & 0.00 & 15629.5 & 26768.7 & 691.4 & 960.1 \\
\cmidrule(lr){2-9} \cmidrule(lr){11-18}
& \multirow{5}{*}{\rotatebox{90}{LNS2}} & 150 & 2.11 & 0.19 & 117.9 & 123.9 & 3.7 & 6.4 & & \multirow{5}{*}{\rotatebox{90}{LNS2}} & 200 & 0.50 & 0.02 & 152.4 & 273.4 & 6.7 & 19.8 \\
 &  & 200 & 0.98 & 0.11 & 247.5 & 261.3 & 8.3 & 11.1 &  &  & 300 & 0.40 & 0.01 & 384.2 & 1976.4 & 24.4 & 106.5 \\
 &  & 250 & 0.70 & 0.07 & 431.7 & 462.3 & 15.3 & 21.9 &  &  & 400 & 0.31 & 0.01 & 1047.9 & 4422.1 & 72.2 & 192.3 \\
 &  & 300 & 0.55 & 0.05 & 694.7 & 757.5 & 26.4 & 41.5 &  &  & 500 & 0.15 & 0.01 & 3671.8 & 10219.3 & 204.6 & 417.5 \\
 &  & 350 & 0.36 & 0.03 & 1042.3 & 1365.1 & 43.4 & 75.2 &  &  & 600 & 0.08 & 0.00 & 11334.0 & 19107.2 & 465.1 & 668.7 \\
\midrule 
\multirow{10}{*}{\rotatebox{90}{den520d}} & \multirow{5}{*}{\rotatebox{90}{LaCAM2}} & 500 & 7.41 & 0.24 & 313.6 & 2545.8 & 28.7 & 174.8 & \multirow{10}{*}{\rotatebox{90}{Paris}} & \multirow{5}{*}{\rotatebox{90}{LaCAM2}} & 350 & 35.13 & 0.70 & 99.3 & \textcolor{red}{98.3} & 4.2 & 22.9 \\
 &  & 600 & 5.34 & 0.14 & 612.7 & 5572.8 & 59.3 & 355.1 &  &  & 450 & 31.77 & 0.48 & 130.4 & 273.6 & 6.3 & 65.3 \\
 &  & 700 & 4.52 & 0.12 & 1110.3 & 8460.1 & 104.3 & 525.7 &  &  & 550 & 25.61 & 0.33 & 204.0 & 1520.5 & 11.2 & 168.1 \\
 &  & 800 & 2.79 & 0.08 & 2410.8 & 16784.3 & 198.8 & 832.8 &  &  & 650 & 22.87 & 0.21 & 267.5 & 4650.5 & 17.2 & 302.8 \\
 &  & 900 & 2.75 & 0.06 & 3648.2 & 27560.2 & 274.3 & 1148.4 &  &  & 750 & 16.96 & 0.15 & 366.1 & 8294.7 & 27.7 & 506.1 \\
\cmidrule(lr){2-9} \cmidrule(lr){11-18}
& \multirow{5}{*}{\rotatebox{90}{LNS2}} & 500 & 6.64 & 0.24 & 296.6 & 2551.8 & 26.0 & 146.3 & & \multirow{5}{*}{\rotatebox{90}{LNS2}} & 350 & 20.20 & 0.71 & 80.4 & 100.9 & 3.0 & 11.6 \\
 &  & 600 & 5.23 & 0.16 & 583.8 & 5301.9 & 52.5 & 241.9 &  &  & 450 & 24.70 & 0.50 & 136.5 & 216.7 & 5.2 & 38.6 \\
 &  & 700 & 4.07 & 0.12 & 1196.9 & 8145.6 & 97.4 & 384.8 &  &  & 550 & 12.81 & 0.33 & 205.3 & 1159.8 & 9.1 & 107.8 \\
 &  & 800 & 3.27 & 0.11 & 2076.0 & 12242.9 & 159.8 & 535.8 &  &  & 650 & 20.25 & 0.23 & 280.0 & 3658.0 & 14.1 & 203.4 \\
 &  & 900 & 2.33 & 0.03 & 3979.5 & 23163.4 & 262.4 & 807.2 &  &  & 750 & 9.60 & 0.17 & 414.2 & 6561.7 & 25.5 & 323.9 \\
\bottomrule
\end{tabular}
    
    }
    
\end{table}

\begin{table*}[b!]
    \centering
    \caption{Final delays and AUC (divided by 10k) of different methods with best neighborhood size, evaluated on maps with differing numbers of agents within $300$s. Initial solver is LNS2. Highlighted are the results ranked \textcolor{Red}{first}, and \textcolor{cyan}{second}.}
\label{table:300s-lns2}
\setlength{\tabcolsep}{1.5pt}
\resizebox{0.8\textwidth}{!}{
\begin{tabular}{@{}c|c|rrrrrrrrrrrrrrrrrrrr@{}}
\toprule
\multicolumn{1}{c|}{\multirow{2}{*}{\rotatebox{90}{map}}} & \multicolumn{1}{c|}{\multirow{2}{*}{n}} & \multicolumn{2}{c}{RW} & \multicolumn{2}{c}{INTC} & \multicolumn{2}{c}{RAND} & \multicolumn{2}{c}{ADP} & \multicolumn{2}{c}{RWP} & \multicolumn{2}{c}{SVM} & \multicolumn{2}{c}{NNS} & \multicolumn{2}{c}{Bandit} & \multicolumn{2}{c}{Ori-SVM} & \multicolumn{2}{c}{Ori-NNS} \\ 
\cmidrule(lr){3-4}\cmidrule(lr){5-6}\cmidrule(lr){7-8}\cmidrule(lr){9-10}\cmidrule(lr){11-12}\cmidrule(lr){13-14}\cmidrule(lr){15-16}\cmidrule(lr){17-18}\cmidrule(lr){19-20}\cmidrule(lr){21-22}
 & & Delay & AUC & Delay & AUC & Delay & AUC & Delay & AUC & Delay & AUC & Delay & AUC & Delay & AUC & Delay & AUC & Delay & AUC & Delay & AUC \\ \midrule

\multicolumn{1}{c|}{\multirow{5}{*}{\rotatebox{90}{empty}}} & 300 & \textcolor{cyan}{358.0} & \textcolor{cyan}{12.2} & 406.4 & 13.8 & 435.1 & 15.2 & 369.0 & 12.6 & 369.9 & 12.6 & 381.5 & 15.4 & 724.6 & 26.9 & 386.3 & 13.3 & 395.6 & 18.2 & \textcolor{Red}{325.9} & \textcolor{Red}{12.9} \\
 & 350 & \textcolor{cyan}{750.9} & \textcolor{cyan}{25.6} & 814.6 & 28.6 & 922.4 & 31.5 & 770.1 & 26.5 & 769.0 & 27.7 & 800.7 & 32.9 & 1209.9 & 42.9 & 811.5 & 28.5 & 807.7 & 32.8 & \textcolor{Red}{734.3} & \textcolor{Red}{28.9} \\
 & 400 & \textcolor{Red}{1397.2} & \textcolor{Red}{49.1} & 1513.7 & 53.7 & 1703.2 & 59.6 & \textcolor{cyan}{1418.6} & \textcolor{cyan}{50.0} & 1431.1 & 50.1 & 1588.3 & 64.5 & 1928.5 & 70.0 & 1537.2 & 54.2 & 1640.4 & 63.9 & 1464.8 & 56.8 \\
 & 450 & \textcolor{Red}{2551.0} & \textcolor{Red}{90.5} & 2695.0 & 95.4 & 2908.1 & 101.8 & \textcolor{cyan}{2577.6} & \textcolor{cyan}{90.7} & 2585.3 & 91.1 & 2766.0 & 105.0 & 3136.3 & 112.2 & 2753.7 & 96.5 & 2936.5 & 109.9 & 2876.6 & 107.5 \\
 & 500 & \textcolor{Red}{4050.5} & 145.4 & 4205.4 & 143.9 & 4438.9 & 155.5 & 4093.8 & \textcolor{Red}{140.4} & \textcolor{cyan}{4051.3} & \textcolor{cyan}{143.9} & 5053.2 & 167.1 & 4803.2 & 179.7 & 4318.5 & 150.8 & 4776.3 & 174.7 & 5305.6 & 186.4 \\ \midrule

\multicolumn{1}{c|}{\multirow{5}{*}{\rotatebox{90}{random}}} & 150 & 332.6 & 10.5 & 352.6 & 11.4 & 357.2 & 11.3 & 330.1 & 10.4 & 337.4 & 10.8 & 331.4 & 10.6 & 360.5 & 11.9 & \textcolor{cyan}{330.1} & \textcolor{cyan}{10.3} & 338.8 & 11.0 & \textcolor{Red}{323.3} & \textcolor{Red}{10.4} \\
 & 200 & \textcolor{cyan}{762.0} & \textcolor{cyan}{25.4} & 811.2 & 26.2 & 831.1 & 27.0 & 771.3 & 25.6 & 784.4 & 25.4 & 786.5 & 27.3 & 900.2 & 30.4 & 779.1 & 24.9 & 790.8 & 26.3 & \textcolor{Red}{728.5} & \textcolor{Red}{24.2} \\
 & 250 & \textcolor{Red}{1504.5} & \textcolor{cyan}{50.4} & 1582.6 & 53.2 & 1606.3 & 53.3 & 1527.5 & 50.6 & 1534.4 & 50.9 & 1713.9 & 59.1 & 1683.7 & 57.5 & \textcolor{cyan}{1507.3} & \textcolor{Red}{49.3} & 1662.8 & 58.3 & 1528.0 & 54.8 \\
 & 300 & \textcolor{Red}{2687.7} & \textcolor{Red}{93.8} & 2839.7 & 98.6 & 2885.9 & 100.1 & \textcolor{cyan}{2737.2} & \textcolor{cyan}{93.9} & 2777.2 & 94.6 & 2885.9 & 104.9 & 3045.2 & 108.4 & 2746.0 & 92.0 & 3005.0 & 107.0 & 3207.8 & 115.5 \\
 & 350 & 4439.2 & \textcolor{cyan}{150.4} & 4609.3 & 155.5 & 4635.2 & 167.8 & \textcolor{cyan}{4432.8} & \textcolor{Red}{150.1} & \textcolor{Red}{4408.2} & 156.6 & 4800.2 & 175.8 & 5466.7 & 193.5 & 4564.1 & 155.8 & 5105.2 & 186.3 & 6004.1 & 206.3 \\ \midrule

\multicolumn{1}{c|}{\multirow{5}{*}{\rotatebox{90}{warehouse}}} & 150 & \textcolor{Red}{117.2} & \textcolor{cyan}{3.7} & 123.5 & 7.0 & 122.6 & 4.8 & \textcolor{cyan}{109.0} & \textcolor{Red}{3.6} & 113.0 & 3.6 & 114.1 & 4.6 & 244.7 & 8.9 & 107.9 & \textcolor{Red}{3.6} & 112.7 & 4.0 & 164.7 & 5.5 \\
 & 200 & \textcolor{cyan}{244.2} & \textcolor{cyan}{8.1} & 317.7 & 15.6 & 286.6 & 12.2 & 242.8 & 8.4 & 252.1 & 8.7 & 245.3 & 9.7 & 469.6 & 17.7 & \textcolor{Red}{239.4} & \textcolor{Red}{8.0} & 252.6 & 9.8 & 269.1 & 10.5 \\
 & 250 & 433.1 & \textcolor{cyan}{15.3} & 695.2 & 32.8 & 537.7 & 23.4 & 435.6 & 16.2 & 443.8 & 16.4 & 439.6 & 19.3 & 749.0 & 31.4 & \textcolor{Red}{414.2} & \textcolor{Red}{14.1} & 441.0 & 21.1 & \textcolor{cyan}{417.4} & 17.7 \\
 & 300 & \textcolor{Red}{663.9} & \textcolor{cyan}{25.6} & 1161.5 & 53.7 & 950.3 & 42.0 & 683.7 & 29.1 & 729.3 & 27.9 & 681.3 & 34.3 & 1501.2 & 57.3 & 669.5 & \textcolor{Red}{23.8} & 703.2 & 36.8 & \textcolor{cyan}{664.8} & 31.1 \\
 & 350 & \textcolor{Red}{1041.8} & \textcolor{cyan}{41.3} & 1949.2 & 84.6 & 1515.9 & 66.2 & 1073.0 & 44.1 & 1134.9 & 44.7 & 1107.4 & 65.7 & 1871.1 & 77.1 & \textcolor{cyan}{1047.6} & \textcolor{Red}{38.1} & 1100.9 & 64.6 & 1094.5 & 56.1 \\ \midrule

\multicolumn{1}{c|}{\multirow{5}{*}{\rotatebox{90}{ost003d}}} & 200 & 150.8 & 6.8 & 290.7 & 14.4 & 194.0 & 9.9 & 149.2 & 6.3 & \textcolor{Red}{147.6} & \textcolor{Red}{5.5} & \textcolor{cyan}{148.6} & \textcolor{cyan}{6.6} & 165.0 & 6.6 & 158.2 & 8.3 & 155.0 & 9.9 & 245.4 & 12.0 \\
 & 300 & 327.3 & 19.9 & 929.5 & 45.6 & 632.0 & 36.5 & 337.6 & 21.3 & \textcolor{Red}{298.0} & \textcolor{Red}{13.9} & \textcolor{cyan}{315.4} & \textcolor{cyan}{23.9} & 338.4 & 15.8 & 532.8 & 33.4 & 338.2 & 34.3 & 582.6 & 39.2 \\
 & 400 & 761.8 & 48.3 & 2048.3 & 100.6 & 1668.8 & 86.3 & 746.1 & 51.0 & \textcolor{Red}{652.3} & \textcolor{Red}{39.3} & 850.3 & 84.3 & \textcolor{cyan}{679.5} & \textcolor{cyan}{41.2} & 1276.3 & 75.2 & 1086.3 & 106.5 & 1483.2 & 109.0 \\
 & 500 & 2051.4 & 142.9 & 4460.2 & 223.1 & 3611.6 & 193.7 & 2094.8 & 141.0 & \textcolor{Red}{1515.2} & \textcolor{Red}{103.4} & 2928.2 & 217.0 & \textcolor{cyan}{2012.2} & \textcolor{cyan}{121.8} & 3059.8 & 163.6 & 3092.7 & 223.4 & 4673.8 & 270.5 \\
 & 600 & 6069.5 & 318.3 & 8824.9 & 404.8 & 8424.3 & 399.2 & 6325.9 & 323.0 & \textcolor{Red}{4731.3} & \textcolor{Red}{260.7} & 10104.8 & 459.1 & \textcolor{cyan}{5501.2} & 294.6 & 6093.6 & \textcolor{cyan}{311.4} & 8604.1 & 457.2 & 11426.4 & 517.9 \\ \midrule

\multicolumn{1}{c|}{\multirow{5}{*}{\rotatebox{90}{den520d}}} & 500 & 293.5 & 22.8 & 1414.3 & 72.4 & 906.1 & 57.9 & 306.2 & 26.1 & \textcolor{Red}{248.5} & \textcolor{Red}{14.1} & \textcolor{cyan}{252.5} & \textcolor{cyan}{28.0} & 313.5 & 20.4 & 607.8 & 47.8 & 394.3 & 60.7 & 732.3 & 87.8 \\
 & 600 & 536.9 & 44.4 & 2293.2 & 116.0 & 1656.1 & 101.5 & 583.1 & 50.2 & \textcolor{cyan}{396.7} & \textcolor{cyan}{26.5} & \textcolor{Red}{389.0} & \textcolor{Red}{46.3} & 496.1 & 37.0 & 1247.0 & 89.2 & 972.6 & 117.2 & 1786.5 & 173.4 \\
 & 700 & 934.1 & 80.3 & 3415.0 & 175.9 & 2907.4 & 171.7 & 1049.8 & 88.0 & \textcolor{Red}{620.2} & \textcolor{Red}{48.3} & \textcolor{cyan}{665.2} & 77.9 & 814.0 & \textcolor{cyan}{68.2} & 2297.3 & 150.2 & 1880.1 & 200.1 & 4228.7 & 286.7 \\
 & 800 & 1476.1 & 123.0 & 4535.4 & 234.8 & 4639.0 & 258.2 & 1685.4 & 131.9 & \textcolor{Red}{883.2} & \textcolor{Red}{75.4} & \textcolor{cyan}{1003.4} & \textcolor{cyan}{142.8} & 1102.0 & 99.7 & 3330.2 & 209.4 & 3876.9 & 312.9 & 7766.6 & 431.3 \\
 & 900 & 2290.1 & 166.7 & 6776.0 & 347.7 & 6447.9 & 345.6 & 2611.1 & 187.9 & \textcolor{Red}{1387.2} & \textcolor{Red}{130.6} & \textcolor{cyan}{1816.6} & 216.5 & 1821.4 & \textcolor{cyan}{168.1} & 5343.4 & 308.8 & 6161.2 & 420.2 & 13367.1 & 624.8 \\ \midrule

\multicolumn{1}{c|}{\multirow{5}{*}{\rotatebox{90}{Paris}}} & 350 & 80.0 & \textcolor{cyan}{2.9} & 121.5 & 9.4 & 83.8 & 6.5 & \textcolor{cyan}{75.5} & 3.3 & 76.4 & \textcolor{Red}{2.7} & \textcolor{Red}{71.9} & 6.3 & 78.6 & 10.7 & 1066.5 & 4.6 & 81.7 & 7.4 & 193.5 & 38.9 \\
 & 450 & 127.4 & \textcolor{Red}{4.9} & 273.6 & 22.2 & 172.7 & 17.2 & 124.9 & 6.6 & \textcolor{Red}{118.8} & \textcolor{cyan}{4.5} & \textcolor{cyan}{120.7} & 12.2 & 142.6 & 7.1 & 737.1 & 11.0 & 123.2 & 20.3 & 130.8 & 42.9 \\
 & 550 & 203.6 & \textcolor{cyan}{9.1} & 540.8 & 38.3 & 345.4 & 33.7 & 192.2 & 11.7 & \textcolor{Red}{183.8} & \textcolor{Red}{7.4} & \textcolor{cyan}{184.7} & 16.5 & 217.5 & 12.2 & 205.3 & 20.7 & 196.5 & 37.6 & 721.5 & 67.7 \\
 & 650 & 278.1 & 14.0 & 963.6 & 66.7 & 650.2 & 57.1 & 270.2 & 17.7 & \textcolor{Red}{257.1} & \textcolor{Red}{12.2} & \textcolor{cyan}{262.0} & \textcolor{cyan}{35.7} & 313.1 & 18.7 & 1022.0 & 37.0 & 286.8 & 78.2 & 307.5 & 128.7 \\
 & 750 & 404.7 & \textcolor{cyan}{25.4} & 1680.8 & 110.9 & 1211.7 & 99.1 & 396.9 & 33.7 & \textcolor{Red}{375.6} & \textcolor{Red}{21.8} & \textcolor{cyan}{388.7} & \textcolor{cyan}{65.3} & 468.2 & 36.6 & 576.9 & 70.6 & 483.2 & 126.6 & 2409.4 & 243.5 \\ \bottomrule
\end{tabular}
}
\end{table*}

\begin{table*}[b!]
    \centering
    \caption{Final delays and AUC (divided by 10k) of different methods with best neighborhood size, evaluated on maps with differing numbers of agents within $60$s. Initial solver is LNS2. Highlighted are the results ranked \textcolor{Red}{first}, and \textcolor{cyan}{second}.}
    \label{table:60s-lns2}
    \setlength{\tabcolsep}{1.5pt}
    \resizebox{0.8\textwidth}{!}{
    \begin{tabular}{@{}c|c|rrrrrrrrrrrrrrrrrrrr@{}}
    \toprule
    \multicolumn{1}{c|}{\multirow{2}{*}{\rotatebox{90}{map}}} & \multicolumn{1}{c|}{\multirow{2}{*}{n}} & \multicolumn{2}{c}{RW} & \multicolumn{2}{c}{INTC} & \multicolumn{2}{c}{RAND} & \multicolumn{2}{c}{ADP} & \multicolumn{2}{c}{RWP} & \multicolumn{2}{c}{SVM} & \multicolumn{2}{c}{NNS} & \multicolumn{2}{c}{Bandit} & \multicolumn{2}{c}{Ori-SVM} & \multicolumn{2}{c}{Ori-NNS} \\ 
    \cmidrule(lr){3-4}\cmidrule(lr){5-6}\cmidrule(lr){7-8}\cmidrule(lr){9-10}\cmidrule(lr){11-12}\cmidrule(lr){13-14}\cmidrule(lr){15-16}\cmidrule(lr){17-18}\cmidrule(lr){19-20}\cmidrule(lr){21-22}
     & & Delay & AUC & Delay & AUC & Delay & AUC & Delay & AUC & Delay & AUC & Delay & AUC & Delay & AUC & Delay & AUC & Delay & AUC & Delay & AUC \\ 
    \midrule \multicolumn{1}{c|}{\multirow{5}{*}{\rotatebox{90}{empty}}} & 300 &\textcolor{Red}{409.2}&\textcolor{Red}{3.0} &463.1&3.5 &522.2&4.0 &\textcolor{cyan}{416.6}&\textcolor{cyan}{3.1} &424.7&\textcolor{cyan}{3.1} &497.3&5.6 &972.6&7.4 &456.7&3.5 &584.5&7.6 &432.3&4.4 \\ 
 & 350 &\textcolor{Red}{885.6}&\textcolor{Red}{6.6} &989.7&7.6 &1083.0&8.2 &\textcolor{cyan}{911.8}&\textcolor{cyan}{6.9} &913.5&\textcolor{cyan}{6.9} &1287.2&11.0 &1540.9&11.3 &994.7&7.7 &1163.0&10.9 &1031.4&9.2 \\ 
 & 400 &\textcolor{Red}{1726.2}&\textcolor{Red}{13.2} &1900.5&14.3 &2105.9&15.2 &1768.8&\textcolor{cyan}{13.3} &\textcolor{cyan}{1765.6}&13.4 &2584.0&19.7 &2461.5&18.2 &1916.4&14.3 &2416.5&18.9 &2127.1&17.3 \\ 
 & 450 &\textcolor{Red}{3198.8}&\textcolor{Red}{22.9} &3359.8&23.9 &3640.0&24.9 &3248.1&\textcolor{cyan}{23.1} &\textcolor{cyan}{3226.7}&23.2 &4094.3&28.4 &4170.4&29.0 &3455.3&24.8 &4148.1&29.6 &4095.3&29.4 \\ 
 & 500 &\textcolor{Red}{4949.4}&\textcolor{Red}{34.9} &5117.5&35.9 &5376.6&36.2 &\textcolor{cyan}{4997.9}&\textcolor{cyan}{35.0} &5024.8&35.5 &6565.9&43.7 &6355.9&43.2 &5473.0&38.0 &6566.5&44.6 &6908.0&46.0 \\ 
\midrule \multicolumn{1}{c|}{\multirow{5}{*}{\rotatebox{90}{random}}} & 150 &354.8&2.4 &374.0&2.5 &379.4&2.5 &\textcolor{cyan}{349.6}&\textcolor{cyan}{2.3} &354.9&\textcolor{cyan}{2.3} &355.8&2.5 &408.2&2.9 &\textcolor{Red}{345.1}&\textcolor{Red}{2.2} &361.1&2.6 &349.8&2.5 \\ 
 & 200 &846.4&\textcolor{cyan}{5.8} &897.1&6.2 &927.2&6.4 &850.0&\textcolor{cyan}{5.8} &868.6&5.9 &963.7&7.3 &1065.9&7.6 &\textcolor{cyan}{845.1}&\textcolor{Red}{5.6} &894.1&6.5 &\textcolor{Red}{830.1}&6.1 \\ 
 & 250 &\textcolor{cyan}{1744.9}&12.5 &1862.1&13.2 &1841.5&13.1 &1762.5&\textcolor{cyan}{12.3} &1761.4&12.5 &1990.0&14.8 &2049.4&14.6 &\textcolor{Red}{1699.9}&\textcolor{Red}{11.7} &2176.5&14.8 &1973.9&14.6 \\ 
 & 300 &\textcolor{Red}{3152.5}&\textcolor{Red}{22.3} &3307.9&23.0 &3351.4&22.9 &3216.7&\textcolor{Red}{22.3} &3235.6&22.6 &4001.1&28.1 &3978.8&27.9 &\textcolor{cyan}{3216.5}&22.6 &4043.2&27.5 &4298.4&29.4 \\ 
 & 350 &\textcolor{cyan}{5280.3}&37.7 &5456.2&38.4 &5416.5&37.4 &5290.9&\textcolor{Red}{37.1} &\textcolor{Red}{5224.1}&\textcolor{Red}{37.1} &6543.7&46.4 &6877.1&46.4 &5540.1&38.5 &7113.9&47.7 &7514.5&48.9 \\ 
\midrule \multicolumn{1}{c|}{\multirow{5}{*}{\rotatebox{90}{warehouse}}} & 150 &121.9&\textcolor{Red}{0.9} &239.7&2.7 &139.4&1.4 &\textcolor{Red}{115.9}&\textcolor{Red}{0.9} &118.7&\textcolor{Red}{0.9} &119.0&1.8 &323.3&2.7 &\textcolor{cyan}{118.2}&\textcolor{Red}{0.9} &126.5&1.2 &178.4&1.5 \\ 
 & 200 &\textcolor{cyan}{258.5}&\textcolor{Red}{2.1} &620.8&6.0 &375.2&4.0 &265.2&2.4 &285.1&2.4 &275.5&3.7 &638.8&5.5 &\textcolor{Red}{256.8}&\textcolor{Red}{2.1} &290.9&3.5 &316.2&3.8 \\ 
 & 250 &\textcolor{cyan}{472.7}&\textcolor{cyan}{4.5} &1333.8&11.7 &791.7&9.1 &488.3&5.2 &534.6&5.1 &600.5&8.0 &1018.6&9.4 &\textcolor{Red}{446.8}&\textcolor{Red}{4.0} &657.2&10.0 &531.7&7.0 \\ 
 & 300 &\textcolor{cyan}{772.8}&\textcolor{cyan}{8.2} &2188.3&18.4 &1510.1&15.5 &839.2&8.8 &888.6&8.3 &1252.4&16.1 &2149.1&18.0 &\textcolor{Red}{758.0}&\textcolor{Red}{7.1} &1265.9&18.2 &1024.5&13.4 \\ 
 & 350 &\textcolor{cyan}{1331.4}&\textcolor{cyan}{14.1} &3346.2&26.9 &2420.4&23.5 &1463.4&15.4 &1455.6&14.3 &3105.0&31.6 &2677.8&22.7 &\textcolor{Red}{1216.4}&\textcolor{Red}{12.0} &2971.2&30.5 &2182.7&24.1 \\ 
\midrule \multicolumn{1}{c|}{\multirow{5}{*}{\rotatebox{90}{ost003d}}} & 200 &\textcolor{cyan}{170.3}&2.6 &522.0&5.6 &331.6&4.4 &176.1&2.5 &\textcolor{Red}{161.4}&\textcolor{Red}{1.5} &200.4&2.8 &205.6&\textcolor{cyan}{2.4} &274.0&4.0 &221.0&5.9 &359.7&5.4 \\ 
 & 300 &517.9&8.3 &1761.0&17.5 &1397.5&15.6 &636.3&9.3 &\textcolor{Red}{386.5}&\textcolor{Red}{5.4} &871.3&13.8 &\textcolor{cyan}{443.0}&\textcolor{cyan}{7.0} &1309.4&15.3 &1543.0&20.6 &1520.9&20.2 \\ 
 & 400 &1785.4&24.3 &3975.2&38.2 &3482.2&34.5 &2013.8&25.3 &\textcolor{Red}{1127.0}&\textcolor{Red}{16.9} &4096.8&44.1 &\textcolor{cyan}{1338.2}&\textcolor{cyan}{19.5} &3127.5&31.4 &5713.4&53.8 &5246.7&49.7 \\ 
 & 500 &5433.1&55.8 &8847.5&75.9 &8225.8&73.5 &6074.1&58.4 &\textcolor{Red}{4083.8}&\textcolor{Red}{45.4} &10580.2&82.1 &\textcolor{cyan}{5033.3}&\textcolor{cyan}{53.8} &6941.8&62.2 &10876.4&90.5 &12373.4&94.0 \\ 
 & 600 &13845.4&110.0 &16830.7&127.1 &16765.7&127.9 &14271.6&108.2 &\textcolor{Red}{11123.2}&\textcolor{Red}{99.2} &18765.4&130.6 &\textcolor{cyan}{12267.2}&\textcolor{cyan}{102.9} &13217.6&106.2 &20754.1&140.7 &21677.9&144.8 \\ 
\midrule \multicolumn{1}{c|}{\multirow{5}{*}{\rotatebox{90}{den520d}}} & 500 &578.3&\textcolor{cyan}{11.2} &2824.2&28.6 &2318.1&26.3 &740.0&13.1 &\textcolor{Red}{335.4}&\textcolor{Red}{7.7} &760.6&20.5 &\textcolor{cyan}{536.7}&11.6 &1947.7&24.2 &3239.3&38.5 &4594.9&46.3 \\ 
 & 600 &1330.7&22.5 &4479.9&45.2 &4246.6&44.2 &1683.4&25.9 &\textcolor{Red}{675.0}&\textcolor{Red}{15.5} &1324.7&33.8 &\textcolor{cyan}{1101.3}&\textcolor{cyan}{22.2} &3808.8&41.3 &6340.4&63.2 &9197.9&74.7 \\ 
 & 700 &2583.3&38.4 &7100.6&68.3 &7256.5&67.9 &3317.7&43.1 &\textcolor{Red}{1498.7}&\textcolor{Red}{28.9} &3562.8&46.6 &\textcolor{cyan}{2115.6}&\textcolor{cyan}{36.8} &6821.9&63.3 &11421.6&93.5 &13704.0&102.6 \\ 
 & 800 &4658.6&60.1 &9677.4&90.0 &11136.0&97.0 &5919.9&66.9 &\textcolor{Red}{2665.5}&\textcolor{Red}{45.7} &7819.7&83.1 &\textcolor{cyan}{3867.7}&\textcolor{cyan}{58.7} &9650.5&85.3 &16086.6&125.6 &19454.3&136.2 \\ 
 & 900 &7159.6&\textcolor{cyan}{84.2} &14664.4&127.0 &14546.4&124.7 &8506.7&89.4 &\textcolor{Red}{4547.5}&\textcolor{Red}{67.6} &10998.2&111.5 &\textcolor{cyan}{6444.6}&85.6 &13552.1&115.6 &20674.6&153.4 &26155.1&172.3 \\ 
\midrule \multicolumn{1}{c|}{\multirow{5}{*}{\rotatebox{90}{Paris}}} & 350 &82.3&\textcolor{Red}{0.9} &332.1&5.1 &176.4&4.0 &\textcolor{cyan}{81.2}&1.5 &\textcolor{Red}{79.6}&\textcolor{Red}{0.9} &95.5&4.3 &305.5&5.2 &82.5&2.9 &84.1&5.4 &1232.7&12.2 \\ 
 & 450 &133.8&\textcolor{cyan}{1.8} &857.4&12.0 &680.5&10.1 &\textcolor{cyan}{129.7}&2.7 &\textcolor{Red}{124.7}&\textcolor{Red}{1.6} &166.9&9.2 &186.9&3.4 &253.2&7.4 &436.5&17.0 &1425.0&21.6 \\ 
 & 550 &\textcolor{cyan}{216.2}&\textcolor{cyan}{4.1} &1494.2&19.8 &1485.0&18.7 &236.4&5.2 &\textcolor{Red}{200.6}&\textcolor{Red}{2.9} &243.5&11.8 &292.3&5.8 &692.2&13.9 &1628.3&29.6 &3342.0&39.5 \\ 
 & 650 &\textcolor{cyan}{302.1}&\textcolor{cyan}{7.1} &2671.8&32.3 &2501.0&28.9 &356.3&9.0 &\textcolor{Red}{291.2}&\textcolor{Red}{5.2} &1081.9&27.6 &447.8&10.4 &1508.9&23.8 &3940.2&59.4 &7253.6&65.1 \\ 
 & 750 &\textcolor{cyan}{559.5}&\textcolor{cyan}{14.5} &4604.3&49.8 &4620.3&45.2 &689.1&16.8 &\textcolor{Red}{549.0}&\textcolor{Red}{10.7} &2267.7&50.0 &907.1&20.6 &3220.9&41.1 &7769.9&82.1 &12831.9&97.3 \\ 
\bottomrule
    \end{tabular}
    }
\end{table*}

\begin{table*}[b!]
    \centering
    \caption{Final delays and AUC (divided by 10k) of different methods with best neighborhood size, evaluated on maps with differing numbers of agents within $300$s. Initial solver is LaCAM2. Highlighted are the results ranked \textcolor{Red}{first}, and \textcolor{cyan}{second}.}
    \label{table:300s-lacam2}
    \setlength{\tabcolsep}{1.5pt}
    \resizebox{0.8\textwidth}{!}{
    \begin{tabular}{@{}c|c|rrrrrrrrrrrrrrrrrrrr@{}}
    \toprule
    \multicolumn{1}{c|}{\multirow{2}{*}{\rotatebox{90}{map}}} & \multicolumn{1}{c|}{\multirow{2}{*}{n}} & \multicolumn{2}{c}{RW} & \multicolumn{2}{c}{INTC} & \multicolumn{2}{c}{RAND} & \multicolumn{2}{c}{ADP} & \multicolumn{2}{c}{RWP} & \multicolumn{2}{c}{SVM} & \multicolumn{2}{c}{NNS} & \multicolumn{2}{c}{Bandit} & \multicolumn{2}{c}{Ori-SVM} & \multicolumn{2}{c}{Ori-NNS} \\ 
    \cmidrule(lr){3-4}\cmidrule(lr){5-6}\cmidrule(lr){7-8}\cmidrule(lr){9-10}\cmidrule(lr){11-12}\cmidrule(lr){13-14}\cmidrule(lr){15-16}\cmidrule(lr){17-18}\cmidrule(lr){19-20}\cmidrule(lr){21-22}
     & & Delay & AUC & Delay & AUC & Delay & AUC & Delay & AUC & Delay & AUC & Delay & AUC & Delay & AUC & Delay & AUC & Delay & AUC & Delay & AUC \\ \midrule

    \multicolumn{1}{c|}{\multirow{5}{*}{\rotatebox{90}{empty}}} & 300 & \textcolor{Red}{350.5} & \textcolor{Red}{12.3} & 402.1 & 14.0 & 441.9 & 15.8 & 369.0 & \textcolor{cyan}{13.0} & \textcolor{cyan}{368.1} & 13.0 & 374.8 & 16.4 & 790.5 & 30.7 & 389.3 & 13.9 & 397.0 & 25.0 & 371.0 & 18.9 \\
    & 350 & \textcolor{Red}{740.3} & \textcolor{Red}{25.9} & 824.2 & 29.1 & 930.2 & 32.8 & \textcolor{cyan}{758.3} & \textcolor{cyan}{27.0} & 759.4 & 27.0 & 823.0 & 40.5 & 1296.4 & 48.7 & 844.4 & 30.2 & 880.2 & 52.6 & 826.4 & 40.4 \\
    & 400 & \textcolor{Red}{1447.9} & \textcolor{Red}{52.9} & 1593.4 & 58.3 & 1756.2 & 61.4 & \textcolor{cyan}{1476.5} & \textcolor{cyan}{53.1} & 1484.4 & 53.9 & 1692.4 & 83.4 & 2094.8 & 81.6 & 1579.0 & 57.6 & 1857.5 & 97.6 & 1614.3 & 74.4 \\
    & 450 & \textcolor{Red}{2638.0} & \textcolor{Red}{96.2} & 2779.5 & 100.9 & 3040.7 & 108.2 & 2684.1 & \textcolor{cyan}{97.2} & \textcolor{cyan}{2681.5} & 98.4 & 3020.0 & 140.3 & 3598.3 & 138.1 & 2828.1 & 103.1 & 3506.0 & 166.1 & 3318.3 & 143.9 \\
    & 500 & \textcolor{Red}{4237.7} & \textcolor{Red}{155.7} & 4448.5 & 162.9 & 4828.2 & 171.8 & 4374.5 & \textcolor{cyan}{156.5} & \textcolor{cyan}{4335.7} & 159.8 & 5194.9 & 223.5 & 5808.0 & 221.1 & 4772.6 & 173.8 & 5793.6 & 244.0 & 6085.3 & 242.9 \\ \midrule

    \multicolumn{1}{c|}{\multirow{5}{*}{\rotatebox{90}{random}}} & 150 & 327.4 & 10.4 & 344.3 & 11.4 & 348.4 & 11.2 & \textcolor{Red}{324.6} & \textcolor{Red}{10.5} & 331.0 & 10.7 & 332.9 & 11.7 & 361.7 & 12.1 & \textcolor{cyan}{326.5} & \textcolor{cyan}{10.3} & 337.6 & 12.3 & 330.2 & 11.9 \\
    & 200 & \textcolor{cyan}{751.1} & \textcolor{cyan}{25.5} & 805.8 & 26.4 & 836.8 & 27.5 & 770.6 & 25.2 & 767.8 & 26.1 & 780.8 & 32.9 & 887.4 & 31.2 & \textcolor{Red}{732.4} & \textcolor{Red}{23.8} & 777.5 & 30.5 & 770.6 & 30.9 \\
    & 250 & \textcolor{cyan}{1472.7} & \textcolor{cyan}{50.2} & 1537.2 & 53.6 & 1613.5 & 54.3 & 1506.7 & 51.0 & 1494.1 & 50.6 & 1521.2 & 60.1 & 1747.1 & 63.6 & \textcolor{Red}{1456.3} & \textcolor{Red}{48.6} & 1604.8 & 67.4 & 1698.5 & 70.9 \\
    & 300 & \textcolor{Red}{2547.4} & \textcolor{Red}{91.9} & 2769.2 & 99.7 & 2811.8 & 100.1 & \textcolor{cyan}{2622.6} & \textcolor{cyan}{92.8} & 2663.6 & 95.3 & 2800.1 & 119.8 & 3171.0 & 122.6 & 2674.4 & 92.3 & 3084.9 & 138.8 & 3435.9 & 147.2 \\
    & 350 & \textcolor{Red}{4341.6} & \textcolor{Red}{169.1} & 4622.2 & 184.6 & 4683.9 & 186.6 & 4406.5 & 163.8 & \textcolor{cyan}{4367.5} & \textcolor{cyan}{166.6} & 5162.7 & 241.2 & 5886.5 & 241.8 & 4598.5 & 169.3 & 5983.6 & 262.9 & 7438.8 & 312.2 \\ \midrule

    \multicolumn{1}{c|}{\multirow{5}{*}{\rotatebox{90}{warehouse}}} & 150 & 114.0 & 4.3 & 127.6 & 7.6 & 121.3 & 5.5 & \textcolor{cyan}{111.4} & 4.2 & 112.8 & \textcolor{Red}{4.2} & 112.0 & 8.9 & 188.7 & 10.2 & \textcolor{Red}{110.3} & \textcolor{cyan}{4.2} & 114.8 & 8.0 & 206.5 & 22.8 \\
    & 200 & 247.5 & 9.2 & 309.6 & 19.1 & 281.0 & 14.0 & \textcolor{cyan}{244.2} & 10.2 & 249.1 & 10.1 & 248.8 & 16.1 & 420.8 & 21.6 & \textcolor{Red}{239.4} & \textcolor{Red}{9.0} & 250.6 & 18.4 & 375.5 & 57.7 \\
    & 250 & 430.0 & 18.1 & 698.5 & 48.0 & 562.4 & 29.8 & \textcolor{Red}{424.9} & \textcolor{Red}{19.3} & 446.7 & 20.1 & 444.6 & 41.1 & 782.1 & 38.3 & \textcolor{cyan}{426.2} & \textcolor{cyan}{16.6} & 445.1 & 36.3 & 961.5 & 127.2 \\
    & 300 & 680.6 & 34.3 & 1264.3 & 86.3 & 973.9 & 56.0 & \textcolor{Red}{673.7} & \textcolor{cyan}{34.7} & 703.5 & 36.9 & \textcolor{cyan}{673.7} & 69.7 & 1423.9 & 78.9 & 695.9 & \textcolor{Red}{29.4} & 716.2 & 70.5 & 2248.7 & 214.7 \\
    & 350 & \textcolor{cyan}{1043.0} & \textcolor{cyan}{56.8} & 2011.0 & 158.7 & 1509.0 & 93.0 & 1055.0 & 63.1 & 1130.8 & 67.0 & \textcolor{Red}{1042.4} & 100.4 & 1907.0 & 101.4 & 1067.6 & \textcolor{Red}{49.7} & 1097.6 & 112.2 & 5255.1 & 361.6 \\ \midrule

    \multicolumn{1}{c|}{\multirow{5}{*}{\rotatebox{90}{ost003d}}} & 200 & 155.6 & 7.5 & 299.4 & 15.7 & 183.3 & 10.3 & 153.3 & 6.8 & \textcolor{Red}{147.4} & \textcolor{Red}{5.9} & \textcolor{cyan}{147.5} & 8.8 & 162.3 & \textcolor{cyan}{7.0} & 170.1 & 9.5 & 154.7 & 15.0 & 496.8 & 44.8 \\
    & 300 & 333.8 & 18.8 & 868.4 & 43.4 & 611.8 & 35.9 & 333.4 & 19.6 & \textcolor{Red}{296.1} & \textcolor{Red}{15.1} & \textcolor{cyan}{297.9} & 27.6 & 339.3 & \textcolor{cyan}{17.4} & 496.5 & 33.5 & 359.8 & 52.2 & 1698.9 & 143.1 \\
    & 400 & 714.7 & 49.3 & 1849.6 & 97.5 & 1488.2 & 91.6 & 805.2 & 55.2 & \textcolor{Red}{622.9} & \textcolor{Red}{38.9} & 762.5 & 96.4 & \textcolor{cyan}{675.3} & \textcolor{cyan}{46.4} & 1253.3 & 78.4 & 1083.7 & 123.7 & 5277.4 & 310.2 \\
    & 500 & 2341.8 & 191.4 & 4338.9 & 243.2 & 3900.2 & 231.6 & \textcolor{cyan}{2067.5} & \textcolor{cyan}{157.8} & \textcolor{Red}{1637.6} & \textcolor{Red}{137.0} & 4386.8 & 314.1 & 2379.1 & 171.6 & 3023.9 & 176.6 & 3719.0 & 312.9 & 12525.6 & 573.9 \\
    & 600 & 7514.0 & 450.6 & 10394.8 & 529.4 & 12466.1 & 613.5 & 8216.9 & 469.6 & \textcolor{Red}{5875.0} & \textcolor{Red}{383.3} & 13066.6 & 618.0 & 8469.1 & 465.0 & \textcolor{cyan}{7244.6} & \textcolor{cyan}{400.1} & 10153.9 & 625.0 & 25804.5 & 963.0 \\ \midrule

    \multicolumn{1}{c|}{\multirow{5}{*}{\rotatebox{90}{den520d}}} & 500 & 311.1 & 27.6 & 1433.0 & 72.9 & 972.7 & 62.8 & 311.5 & 29.1 & \textcolor{Red}{241.6} & \textcolor{Red}{16.4} & \textcolor{Red}{240.1} & 33.4 & 322.6 & \textcolor{cyan}{26.1} & 764.6 & 64.4 & 440.5 & 102.5 & 3437.9 & 272.1 \\
    & 600 & 546.3 & 52.3 & 2289.3 & 118.9 & 1649.3 & 109.2 & 597.4 & 55.6 & \textcolor{cyan}{386.6} & \textcolor{cyan}{31.2} & \textcolor{Red}{381.0} & 75.9 & 513.4 & 50.2 & 1235.4 & 95.9 & 1353.4 & 228.2 & 8231.3 & 469.5 \\
    & 700 & 933.1 & 87.0 & 3163.0 & 169.0 & 2782.7 & 176.2 & 1075.2 & 91.3 & \textcolor{Red}{622.4} & \textcolor{Red}{55.4} & 660.7 & 138.1 & 816.6 & \textcolor{cyan}{85.9} & 2439.2 & 165.8 & 2259.0 & 327.1 & 15529.0 & 726.0 \\
    & 800 & 1548.6 & 128.4 & 4437.5 & 245.0 & 4337.2 & 266.4 & 1574.1 & 132.9 & \textcolor{Red}{911.5} & \textcolor{Red}{95.2} & \textcolor{cyan}{1003.6} & 200.9 & 1323.8 & 147.8 & 3671.6 & 244.9 & 4906.7 & 499.1 & 21966.5 & 952.2 \\
    & 900 & 2486.1 & \textcolor{cyan}{201.9} & 6426.0 & 351.1 & 6923.5 & 396.9 & 2819.9 & 225.6 & \textcolor{Red}{1382.0} & \textcolor{Red}{151.6} & 2058.6 & 335.4 & 2067.1 & 217.6 & 5595.4 & 346.7 & 15199.0 & 723.9 & 31694.0 & 1247.0 \\ \midrule

    \multicolumn{1}{c|}{\multirow{5}{*}{\rotatebox{90}{Paris}}} & 350 & 87.4 & \textcolor{cyan}{4.0} & 130.6 & 10.4 & 84.4 & 8.3 & \textcolor{cyan}{78.1} & 4.4 & \textcolor{Red}{75.8} & \textcolor{Red}{3.6} & 79.0 & 21.3 & 181.7 & 14.2 & 83.6 & 6.7 & 80.0 & 20.0 & 1285.3 & 78.7 \\
    & 450 & 132.5 & \textcolor{cyan}{6.5} & 274.9 & 22.8 & 174.6 & 19.0 & 123.3 & 7.9 & \textcolor{Red}{118.4} & \textcolor{Red}{5.7} & \textcolor{cyan}{119.7} & 16.0 & 144.4 & 9.5 & 133.4 & 15.0 & 133.0 & 46.6 & 1483.0 & 134.5 \\
    & 550 & 198.0 & \textcolor{cyan}{10.7} & 776.5 & 58.2 & 466.9 & 48.6 & 194.8 & 15.3 & \textcolor{Red}{183.3} & \textcolor{Red}{9.7} & \textcolor{cyan}{184.7} & 33.0 & 228.9 & 17.1 & 210.2 & 28.7 & 200.6 & 105.3 & 2643.7 & 261.4 \\
    & 650 & 276.7 & \textcolor{cyan}{16.9} & 960.1 & 74.0 & 758.5 & 72.3 & 275.4 & 20.7 & \textcolor{Red}{257.5} & \textcolor{Red}{14.1} & \textcolor{cyan}{260.6} & 70.0 & 319.2 & 28.4 & 337.4 & 50.1 & 319.8 & 145.7 & 7697.9 & 459.3 \\
    & 750 & 373.2 & \textcolor{cyan}{27.5} & 1821.2 & 123.2 & 1362.6 & 115.0 & 385.8 & 39.4 & \textcolor{Red}{366.0} & \textcolor{Red}{23.9} & \textcolor{cyan}{369.9} & 108.0 & 498.8 & 47.8 & 627.3 & 87.1 & 616.3 & 261.5 & 13831.5 & 656.4 \\ \bottomrule
    \end{tabular}
    }
\end{table*}

\begin{table*}[b!]
    \centering
    \caption{Final delays and AUC (divided by 10k) of different methods with best neighborhood size, evaluated on maps with differing numbers of agents within $60$s. Initial solver is LaCAM2. Highlighted are the results ranked \textcolor{Red}{first}, and \textcolor{cyan}{second}.}
    \label{table:60s-lacam2}
    \setlength{\tabcolsep}{1.5pt}
    \resizebox{0.8\textwidth}{!}{
    \begin{tabular}{@{}c|c|rrrrrrrrrrrrrrrrrrrr@{}}
    \toprule
    \multicolumn{1}{c|}{\multirow{2}{*}{\rotatebox{90}{map}}} & \multicolumn{1}{c|}{\multirow{2}{*}{n}} & \multicolumn{2}{c}{RW} & \multicolumn{2}{c}{INTC} & \multicolumn{2}{c}{RAND} & \multicolumn{2}{c}{ADP} & \multicolumn{2}{c}{RWP} & \multicolumn{2}{c}{SVM} & \multicolumn{2}{c}{NNS} & \multicolumn{2}{c}{Bandit} & \multicolumn{2}{c}{Ori-SVM} & \multicolumn{2}{c}{Ori-NNS} \\ 
    \cmidrule(lr){3-4}\cmidrule(lr){5-6}\cmidrule(lr){7-8}\cmidrule(lr){9-10}\cmidrule(lr){11-12}\cmidrule(lr){13-14}\cmidrule(lr){15-16}\cmidrule(lr){17-18}\cmidrule(lr){19-20}\cmidrule(lr){21-22}
     & & Delay & AUC & Delay & AUC & Delay & AUC & Delay & AUC & Delay & AUC & Delay & AUC & Delay & AUC & Delay & AUC & Delay & AUC & Delay & AUC \\ 
    \midrule \multicolumn{1}{c|}{\multirow{5}{*}{\rotatebox{90}{empty}}} & 300 &\textcolor{Red}{402.2}&\textcolor{Red}{3.5} &461.9&3.9 &529.8&4.3 &421.5&\textcolor{Red}{3.5} &\textcolor{cyan}{420.6}&\textcolor{Red}{3.5} &513.9&6.7 &1111.1&9.2 &465.4&4.0 &1065.8&13.2 &584.3&8.7 \\ 
 & 350 &\textcolor{Red}{869.9}&\textcolor{Red}{7.2} &994.5&8.2 &1115.7&8.9 &\textcolor{cyan}{917.5}&7.6 &919.0&\textcolor{cyan}{7.5} &1814.8&16.6 &1741.3&14.1 &1041.3&8.4 &2497.8&23.1 &1449.0&16.8 \\ 
 & 400 &\textcolor{Red}{1845.7}&15.2 &2030.1&16.6 &2130.5&16.4 &\textcolor{cyan}{1861.8}&\textcolor{Red}{14.8} &1877.7&\textcolor{cyan}{15.1} &3679.9&30.5 &2981.6&24.6 &2027.8&16.1 &4503.6&34.8 &2942.8&27.9 \\ 
 & 450 &\textcolor{Red}{3414.6}&27.3 &3581.4&28.2 &3827.3&29.1 &\textcolor{cyan}{3455.5}&\textcolor{Red}{26.9} &3513.2&\textcolor{Red}{26.9} &5998.8&45.5 &5185.4&39.6 &3679.1&28.9 &7318.2&51.7 &5961.3&46.3 \\ 
 & 500 &\textcolor{Red}{5605.0}&43.8 &5895.3&45.6 &6076.0&\textcolor{cyan}{43.2} &\textcolor{cyan}{5631.0}&\textcolor{Red}{41.8} &5790.6&44.7 &8973.6&63.7 &8475.4&60.3 &6294.5&48.1 &9866.2&67.5 &9953.8&68.3 \\ 
\midrule \multicolumn{1}{c|}{\multirow{5}{*}{\rotatebox{90}{random}}} & 150 &\textcolor{cyan}{344.7}&\textcolor{cyan}{2.5} &367.3&2.6 &373.6&2.6 &348.4&\textcolor{cyan}{2.5} &355.1&2.6 &357.2&3.5 &409.5&3.1 &\textcolor{Red}{342.4}&\textcolor{Red}{2.4} &375.2&4.0 &386.2&3.6 \\ 
 & 200 &\textcolor{cyan}{840.0}&\textcolor{cyan}{6.1} &885.8&6.5 &930.6&6.6 &843.3&\textcolor{cyan}{6.1} &846.5&6.2 &1085.9&12.7 &1093.4&8.4 &\textcolor{Red}{799.1}&\textcolor{Red}{5.7} &996.5&10.7 &1045.8&10.7 \\ 
 & 250 &\textcolor{cyan}{1698.3}&13.2 &1830.5&14.5 &1832.2&14.0 &1733.5&\textcolor{cyan}{13.1} &1711.6&\textcolor{cyan}{13.1} &2274.6&19.9 &2246.3&18.2 &\textcolor{Red}{1652.8}&\textcolor{Red}{12.2} &2587.1&23.6 &2674.5&23.3 \\ 
 & 300 &\textcolor{cyan}{3083.6}&\textcolor{cyan}{23.9} &3254.9&25.6 &3323.0&24.6 &3175.9&25.9 &\textcolor{Red}{3076.8}&\textcolor{Red}{23.8} &4683.8&40.2 &4521.7&36.6 &3149.7&24.7 &5975.6&47.1 &5967.4&46.2 \\ 
 & 350 &\textcolor{cyan}{5535.2}&46.4 &6027.2&50.6 &5764.9&46.7 &5544.0&\textcolor{Red}{45.5} &\textcolor{Red}{5463.2}&\textcolor{cyan}{45.6} &10463.7&75.4 &9710.6&71.7 &5944.0&49.9 &10815.6&76.9 &12978.2&83.9 \\ 
\midrule \multicolumn{1}{c|}{\multirow{5}{*}{\rotatebox{90}{warehouse}}} & 150 &121.5&\textcolor{Red}{1.4} &238.2&3.9 &147.8&2.3 &\textcolor{Red}{116.5}&1.5 &\textcolor{cyan}{116.6}&\textcolor{Red}{1.4} &124.9&6.1 &349.1&4.2 &118.7&1.5 &124.8&5.2 &869.0&15.5 \\ 
 & 200 &\textcolor{Red}{259.1}&\textcolor{Red}{3.1} &635.2&9.7 &368.0&6.1 &\textcolor{cyan}{261.1}&3.9 &288.2&4.2 &301.7&9.9 &722.1&8.9 &270.8&\textcolor{Red}{3.1} &298.4&12.1 &3349.3&33.9 \\ 
 & 250 &\textcolor{cyan}{470.1}&\textcolor{cyan}{7.2} &1731.4&24.9 &813.2&14.3 &485.4&8.7 &537.1&8.8 &1144.1&28.8 &1201.9&16.1 &\textcolor{Red}{466.8}&\textcolor{Red}{6.1} &873.0&24.4 &7424.5&60.5 \\ 
 & 300 &\textcolor{cyan}{837.5}&\textcolor{cyan}{14.8} &3361.8&42.1 &1631.3&28.7 &864.6&17.3 &935.5&18.6 &2498.1&48.7 &2821.1&33.7 &\textcolor{Red}{809.0}&\textcolor{Red}{11.9} &3117.9&46.5 &11665.5&85.4 \\ 
 & 350 &\textcolor{cyan}{1421.0}&\textcolor{cyan}{30.4} &7091.6&72.7 &2764.4&49.8 &1577.1&33.5 &1514.4&34.2 &4450.8&63.3 &3366.5&45.0 &\textcolor{Red}{1334.3}&\textcolor{Red}{22.4} &5331.1&71.1 &17479.6&119.6 \\ 
\midrule \multicolumn{1}{c|}{\multirow{5}{*}{\rotatebox{90}{ost003d}}} & 200 &\textcolor{cyan}{173.5}&3.2 &556.1&6.7 &336.9&5.0 &180.0&2.9 &\textcolor{Red}{161.1}&\textcolor{Red}{2.3} &174.3&5.2 &204.0&\textcolor{cyan}{2.8} &302.0&4.7 &269.6&10.9 &2176.5&23.8 \\ 
 & 300 &547.6&9.7 &1577.8&17.4 &1308.2&16.0 &604.4&10.2 &\textcolor{Red}{376.7}&\textcolor{Red}{6.6} &991.2&18.1 &\textcolor{cyan}{480.7}&\textcolor{cyan}{8.5} &1305.7&16.2 &2078.2&35.1 &7563.8&58.8 \\ 
 & 400 &1717.2&27.2 &3618.8&41.1 &3750.1&41.2 &2023.8&29.1 &\textcolor{Red}{1143.8}&\textcolor{Red}{21.4} &5073.1&52.7 &\textcolor{cyan}{1541.2}&\textcolor{cyan}{25.7} &3118.4&35.2 &6332.8&69.6 &14720.5&102.2 \\ 
 & 500 &7329.4&78.1 &10386.2&96.3 &9934.0&93.0 &\textcolor{cyan}{7268.4}&76.0 &\textcolor{Red}{5425.1}&\textcolor{Red}{66.6} &14676.1&117.5 &7609.4&78.6 &7467.7&\textcolor{cyan}{73.2} &15632.3&125.7 &24107.7&156.0 \\ 
 & 600 &20549.4&163.0 &21653.0&171.5 &22508.1&173.3 &20281.3&160.2 &\textcolor{cyan}{17560.6}&\textcolor{cyan}{153.4} &26121.5&188.0 &20302.9&162.1 &\textcolor{Red}{16993.8}&\textcolor{Red}{146.6} &29866.4&204.3 &36312.5&224.6 \\ 
\midrule \multicolumn{1}{c|}{\multirow{5}{*}{\rotatebox{90}{den520d}}} & 500 &659.4&\textcolor{cyan}{14.5} &2692.4&30.3 &2410.6&30.0 &885.1&17.5 &\textcolor{Red}{324.1}&\textcolor{Red}{10.1} &625.9&26.4 &\textcolor{cyan}{580.2}&17.1 &2792.7&33.0 &5518.3&71.4 &13834.4&99.3 \\ 
 & 600 &1483.9&\textcolor{cyan}{28.8} &4485.7&49.8 &4631.4&50.8 &1630.7&29.2 &\textcolor{Red}{728.0}&\textcolor{Red}{20.2} &3301.9&57.7 &\textcolor{cyan}{1419.7}&33.7 &3979.6&47.5 &14197.2&119.0 &21399.6&141.3 \\ 
 & 700 &\textcolor{cyan}{2642.0}&\textcolor{cyan}{46.8} &6254.7&71.3 &7336.5&76.6 &3418.0&51.8 &\textcolor{Red}{1494.6}&\textcolor{Red}{36.2} &6070.3&97.5 &2934.0&56.2 &7198.2&75.1 &19781.8&158.8 &30196.8&191.7 \\ 
 & 800 &\textcolor{cyan}{4720.1}&\textcolor{cyan}{73.6} &9333.7&101.2 &11107.2&110.0 &5110.1&\textcolor{cyan}{73.6} &\textcolor{Red}{2774.8}&\textcolor{Red}{58.0} &10157.1&128.7 &5846.9&92.4 &10615.0&106.9 &26036.7&198.6 &38304.4&241.5 \\ 
 & 900 &\textcolor{cyan}{8274.1}&\textcolor{cyan}{108.8} &13091.0&138.6 &16636.2&154.7 &9786.3&114.8 &\textcolor{Red}{4788.8}&\textcolor{Red}{86.8} &18143.9&185.3 &9199.7&126.9 &14398.9&144.2 &30208.9&191.0 &47864.2&297.4 \\ 
\midrule \multicolumn{1}{c|}{\multirow{5}{*}{\rotatebox{90}{Paris}}} & 350 &93.3&\textcolor{cyan}{1.8} &318.1&6.1 &227.5&5.6 &\textcolor{cyan}{84.9}&1.9 &\textcolor{Red}{79.1}&\textcolor{Red}{1.6} &228.5&19.2 &407.7&8.2 &111.1&4.6 &171.8&18.0 &3826.4&36.5 \\ 
 & 450 &136.5&\textcolor{cyan}{3.3} &842.0&12.9 &664.6&12.4 &\textcolor{cyan}{128.6}&3.6 &\textcolor{Red}{124.5}&\textcolor{Red}{2.7} &180.9&13.0 &210.9&5.6 &356.3&11.1 &1575.7&41.3 &7541.4&62.5 \\ 
 & 550 &\textcolor{cyan}{217.2}&\textcolor{cyan}{5.9} &2269.4&27.1 &2110.1&27.1 &236.4&7.4 &\textcolor{Red}{202.5}&\textcolor{Red}{5.2} &545.6&27.9 &350.3&10.9 &960.8&20.8 &7313.9&78.0 &13938.8&99.1 \\ 
 & 650 &\textcolor{cyan}{307.5}&\textcolor{cyan}{10.0} &2891.9&37.3 &3069.1&38.2 &380.5&13.8 &\textcolor{Red}{287.6}&\textcolor{Red}{7.7} &2474.7&58.9 &549.9&19.3 &2019.2&33.8 &9380.0&103.1 &20920.9&137.6 \\ 
 & 750 &\textcolor{cyan}{471.4}&\textcolor{cyan}{18.0} &4794.1&57.7 &5078.5&55.7 &681.7&22.3 &\textcolor{Red}{462.8}&\textcolor{Red}{13.6} &5088.3&85.1 &1080.6&32.8 &3743.0&53.0 &18276.6&146.3 &27245.0&175.1 \\ 
\bottomrule
    \end{tabular}
    }
\end{table*}

\subsection{EECBS as Initial Solver}

The final delays when initial solver is EECBS in maps with a medium number of agents are shown in Table~\ref{table:eecbs-as-initial}. 

\begin{table}[ht]
\caption{Final delays across methods for in maps with the medium number of agent. Time limits are $300$s and $60$s, respectively. Initial solver is EECBS.}

\vspace{-8pt}

\label{table:eecbs-as-initial}
\centering
\setlength{\tabcolsep}{0.5pt}
\resizebox{0.46\textwidth}{!}{
\begin{tabular}{l|cccccc|cccccc}
\toprule

\multicolumn{1}{c}{}  & \multicolumn{6}{|c|}{\textbf{Medium Number of Agents; Time: 300s}} & \multicolumn{6}{c}{\textbf{Medium Number of Agents; Time: 60s}}\\
\midrule

\multirow{1}{*}{\textbf{Methods}}  
& \makecell{\textbf{empty} \\ \textbf{+400}} & \makecell{\textbf{random} \\ \textbf{+250}} & \makecell{\textbf{warehouse} \\ \textbf{+250}} & \makecell{\textbf{ost003d} \\ \textbf{+400}} & \makecell{\textbf{den520d} \\ \textbf{+700}} & \makecell{\textbf{Paris} \\ \textbf{+550}} 
& \makecell{\textbf{empty} \\ \textbf{+400}} & \makecell{\textbf{random} \\ \textbf{+250}} & \makecell{\textbf{warehouse} \\ \textbf{+250}} & \makecell{\textbf{ost003d} \\ \textbf{+400}} & \makecell{\textbf{den520d} \\ \textbf{+700}} & \makecell{\textbf{Paris} \\ \textbf{+550}} \\
\midrule
RW &  \textcolor{Red}{1116.1} & - & \textcolor{Red}{402.0} & - & - & - 
& \textcolor{Red}{1313.8} & - & \textcolor{cyan}{436.9} & - & - & - \\
INT & 1228.2 & - & 470.4 & - & - & - 
& 1412.4 & - & 549.7 & - & - & - \\
RAND & 1311.3 & - & 440.4 & - & - & - 
& 1438.8 & - & 474.5 & - & - & - \\
ADP & \textcolor{cyan}{1170.1} & - & 408.8 & - & - & - 
& \textcolor{cyan}{1342.7} & - & 446.6 & - & - & - \\
RWP & 1202.1 & - & 430.5 & - & - & - 
& 1369.4 & - & 488.4 & - & - & - \\
\midrule
SVM & 1252.4 & - & 408.1 & - & - & - 
& 1826.7 & - & 451.0 & - & - & - \\
NNS & 1506.5 & - & 524.0 & - & - & - 
& 1823.5 & - & 599.9 & - & - & - \\
Bandit & 1250.0 & - & \textcolor{cyan}{404.8} & - & - & - 
& 1474.0 & - & \textcolor{Red}{425.4} & - & - & - \\
Ori-SVM & 1313.6 & - & 417.2 & - & - & - 
& 1968.9 & - & 455.8 & - & - & - \\
Ori-NNS & 1409.7 & - & 497.1 & - & - & - 
& 1946.7 & - & 694.5 & - & - & - \\
\bottomrule

\end{tabular}
}

\vspace{3pt}

\parbox{0.46\textwidth}{\scriptsize Note: RW, INT, RAND, ADP, and RWP stand for RandomWalk, Intersection, Random, Adaptive, and RandomWalkProb, respectively. The agent numbers are shown after the name of a map. Highlighted are the results ranked \textcolor{Red}{first}, and \textcolor{cyan}{second}.}

\end{table}

\clearpage

\end{document}